\documentclass[journal]{IEEEtran}
\usepackage{graphicx}
\usepackage{multirow}
\usepackage{longtable}
\usepackage{rotating}
\usepackage{color}
\usepackage{makecell}
\usepackage[pagebackref=true,breaklinks=true,letterpaper=true,colorlinks,bookmarks=false]{hyperref}
\usepackage{cite}
\usepackage{lineno}
\usepackage{amsmath}
\usepackage{amssymb}
\usepackage{makecell}

\graphicspath{{figures/}}

\begin{document}
%%%%%%%%% TITLE
\title{NestFuse: An Infrared and Visible Image Fusion Architecture based on Nest Connection and Spatial/Channel Attention Models}

\author{Hui~Li,
        Xiao-Jun~Wu
	    and Tariq Durrani,~\IEEEmembership{Fellow,~IEEE}
        % <-this % stops a space
\thanks{This work was supported by the National Key Research and Development Program of China (Grant No. 2017YFC1601800), the National Natural Science Foundation of China (61672265, U1836218), the 111 Project of Ministry of Education of China (B12018).}
\thanks{Hui Li and Xiao-Jun~Wu(\emph{Corresponding author}) are with the School of Artificial Intelligence and Computer Science, Jiangnan University, Wuxi 214122, China. (hui\_li\_jnu@163.com, xiaojun\_wu\_jnu@163.com)}
\thanks{Tariq Durrani is with the Department of Electronic and Electrical Engineering, University of Strathclyde, G1 1XW, Glasgow, UK. (e-mail: t.durrani@strath.ac.uk)}
%\thanks{This paper has supplementary downloadable material available at http://ieeexplore.ieee.org., provided by the authors.}
}

\markboth{}%
{Li \MakeLowercase{\textit{et al.}}: NestFuse: An Infrared and Visible Image Fusion Architecture based on Nest Connection and Spatial/Channel Attention Models}

\maketitle
%\thispagestyle{empty}

%%%%%%%%% ABSTRACT
\begin{abstract}
In this paper we propose a novel method for infrared and visible image fusion where we develop nest connection-based network and spatial/channel attention models. The nest connection-based network can preserve significant amounts of information from input data in a multi-scale perspective. The approach comprises three key elements: encoder, fusion strategy and decoder respectively. In our proposed fusion strategy, spatial attention models and channel attention models are developed that describe the importance of each spatial position and of each channel with deep features. Firstly, the source images are fed into the encoder to extract multi-scale deep features. The novel fusion strategy is then developed to fuse these features for each scale. Finally, the fused image is reconstructed by the nest connection-based decoder. Experiments are performed on publicly available datasets. These exhibit that our proposed approach has better fusion performance than other state-of-the-art methods. This claim is justified through both subjective and objective evaluation. The code of our fusion method is available at \url{https://github.com/hli1221/imagefusion-nestfuse}.
\end{abstract}


%%%%%%%%% BODY TEXT
\IEEEpeerreviewmaketitle

\section{Introduction}
\label{intro}

\IEEEPARstart{I}{mage} fusion represents an important technique in image processing aimed at generating a single image containing salient features and complementary information from source images, by using appropriate feature extraction methods and fusion strategies \cite{li2017pixel}. Current state-of-the-art fusion algorithms are widely employed in many applications, such as in self-driving vehicles, visual tracking \cite{li2020mdlatlrr} \cite{li2019rgb} \cite{vot2019rgbt} and video surveillance.

Fusion algorithms can be broadly classified into two categories: traditional methods \cite{pajares2004wavelet} \cite{ben2005multiscale} \cite{li2020laplacian} \cite{liu2017infrared} \cite{li2019discriminative} \cite{li2017multi} and deep learning-based methods \cite{liu2016image} \cite{li2018infrared} \cite{li2019infrared} \cite{li2018densefuse}. Most traditional methods are based on signal processing operators that have achieved good performance. In recent years, deep learning-based methods have exhibited immense potential in image fusion tasks and have been seen to offer better performance than traditional algorithms.

Traditional methods, in general, cover two approaches: multi-scale based methods; sparse and low-rank representation learning-based methods. Multi-scale methods \cite{pajares2004wavelet} \cite{ben2005multiscale} \cite{yang2010image} \cite{li2013image} \cite{vishwakarma2018image} usually decompose source images into different scales to extract features and use appropriate fusion strategies to fuse each scale feature. An inverse operator is then used to reconstruct the fused image. Although these methods demonstrate good fusion performance, their performance is highly dependent on the multi-scale methods.

Before the development of deep learning-based fusion methods, the sparse representation \cite{wright2008robust} (SR) and low-rank representation \cite{liu2010robust} (LRR) had attracted significant attention. Based on SR, several fusion algorithms were developed \cite{liu2017infrared} \cite{li2019discriminative} \cite{singh2019multimodal}. In \cite{liu2017infrared}, Liu et al. proposed a fusion algorithm based on joint sparse representation (JSR) and saliency detection operator. The JSR is used to extract common information and complementary features from source images.

%For complementary features, the saliency detection operator is utilized to recognize salient region and fuse these features by $l_1$-norm based fusion strategy. Finally, the fused image is obtained by combining the common information and the fused salient features.

In LRR domain, Li et al. \cite{li2017multi} presented a multi-focus image fusion method based on LRR and dictionary learning. In this approach, firstly, source images are divided into image patches and the histogram of oriented gradient (HOG) features are utilized to classify each image patch. A global dictionary is learned by K-singular value decomposition (K-SVD) \cite{aharon2006k}. In addition, there are many other methods combining SR and other operators, such as pulse coupled neural network (PCNN) \cite{lu2014infrared}, and the shearlet transform \cite{yin2017novel}.

Although the SR and LRR based fusion methods have indicated very good performance. These methods still have weaknesses: (1) The running time of fusion algorithms is highly dependent on the dictionary learning operator; (2) When source images are complex, this leads to representation performance degradation.

To solve these draw backs, in the past several years, many deep learning-based fusion methods have been proposed. These methods can be separated into two categories: with and without training phase.

\begin{table*}[!ht]
\centering
\caption{\label{tab:intro}The summary of the existing fusion methods.}
\resizebox{\linewidth}{!}{
\begin{tabular}{|c|c|c|c|c|}
\hline
 First class & Second class & Reference & Advantages & Disadvantages \\
\hline
\multirow{2}*{\makecell*[c]{Traditional \\methods}} &
		Multi-scale & \makecell*[l]{Wavelet\cite{pajares2004wavelet}, Biorthogonal \\ wavelet\cite{ben2005multiscale},  Contourlet\cite{yang2010image}, \\ Guided filtering\cite{li2013image}, Non-\\subsampled shearlet \\transform (NSST)\cite{vishwakarma2018image}} & \makecell*[l]{The raw data is transformed into freq-\\uency domain, which may extract more \\ useful information to represent the \\ source images. With the appropriate \\fusion strategies, these methods may \\achieve better performance.} & \makecell*[l]{(1) Their performance is highly dependent on \\ the multi-scale methods, which are complex \\ to find an appropriate decomposition method \\ for different type of source images; \\ (2) In transform processing, it may cause \\ unrecoverable loss of data.} \\
\cline{2-5}
~	&	SR/LRR		& \makecell*[l]{JSRSD\cite{liu2017infrared}, DDL\cite{li2019discriminative}, \\ Sparse K-SVD\cite{singh2019multimodal},  \\ DLLRR\cite{li2017multi}, TS-SR\cite{lu2014infrared}, \\ DCST-SR\cite{yin2017novel}, \\ConvSR\cite{liu2016image}} & \makecell*[l]{Unlike multi-scale transform, the SR \\ and LRR based methods directly do \\ the fusion process without the trans-\\ form processing. Furthermore, these \\ methods can avoid the unrecoverable \\loss of data.} & \makecell*[l]{(1) The running time of fusion algorithms \\ is highly dependent on the dictionary \\ learning operator; \\ (2) When source images are complex, this \\ leads to representation performance \\ degradation.} \\
\hline
\multirow{2}*{\makecell*[c]{Deep learning \\based methods}} &
		\makecell*[c]{Without \\ training phase} & \makecell*[l]{VggML\cite{li2018infrared}, ResNet-ZCA\cite{li2019infrared}}& \makecell*[l]{(1) This is the first time that the pretr-\\ained deep neural networks which \\ can extract multi-level deep features \\ are utilized in image fusion task;\\
(2) The multi-level deep features \\ contains richer information which is \\ benefit for the image fusion tasks.
} & \makecell*[l]{(1) Since these pre-trained networks are \\ trained for different tasks, they may not fit \\ the image fusion tasks;\\ 
(2) The deep feature extraction operation can \\ be improved by train an appropriate fusion \\ network.
} \\
\cline{2-5}
~	&	\makecell*[c]{With \\training phase}	 & \makecell*[l]{CNN\cite{liu2017multi}, Unsupervised\cite{yan2018unsupervised}, \\ DneseFuse\cite{li2018densefuse}, FusionGAN\\ \cite{ma2019fusiongan}, IFCNN\cite{zhang2020ifcnn}} & \makecell*[l]{(1) With appropriate fusion network,\\the deep features contains more useful \\information \cite{liu2017multi}\cite{yan2018unsupervised};\\(2) The auto-encoder based fusion net-\\work \cite{li2018densefuse} avoid the lack of training \\ data in image fusion task;\\ (3) The end-to-end image fusion frame-\\works \cite{ma2019fusiongan} \cite{zhang2020ifcnn} can generate the fused \\image without any handcrafted feature \\extraction operation.
} & \makecell*[l]{(1) The network has no down-sampling \\ operator, which can not extract multi-scale \\features. And the deep features are not fully \\ utilized; \\ (2) The topology of network architecture \\ need to be improved for multi-scale \\ feature extraction; \\ (3) The fusion strategy is not carefully \\ designed for the fusion of deep features.} \\
\hline
\end{tabular}}
\end{table*}

Without the training phase implies that these methods do not have backpropagation, and use a pre-trained network to extract deep features, which leads to the generation of a decision map.  Based on this theory, Li et al. \cite{li2018infrared}\cite{li2019infrared} proposed a fusion framework that utilizes pre-trained network (VGG-19 \cite{simonyan2014very} and ResNet50 \cite{he2016deep}). This was the first time that multi-level deep features were used to address the infrared and visible image fusion task.

%Moreover, Liu et al. \cite{liu2016image} applied convolutional sparse representation (ConvSR) into image fusion field and the deep features were extracted by ConvSR.

Since an appropriate model for image fusion task can be trained to obtain better fusion performance, the latest deep learning methods are all based on this strategy. In 2017, Liu et al. \cite{liu2017multi} proposed a convolutional neural network (CNN) \cite{krizhevsky2012imagenet} used a fusion framework for the multi-focus image fusion task. Yan et al. \cite{yan2018unsupervised} also presented a fusion network based on CNN and multi-level features. In the infrared and visible image fusion field, Li et al. \cite{li2018densefuse} proposed a novel fusion framework based on a dense block \cite{huang2017densely} and an auto-encoder architecture. Ma et al. \cite{ma2019fusiongan} applied generative adversarial network (GAN)\cite{goodfellow2014generative} for the infrared and visible image fusion task. Compared with existing fusion methods, these CNN or GAN based fusion frameworks have achieved extraordinary fusion performance.

However, these deep learning-based frameworks still have several drawbacks: (1) The network has no down-sampling operator and cannot extract multi-scale features, and the deep features are not fully utilized; (2) The topology of network architecture needs to be improved for multi-scale feature extraction; (3) The fusion strategy is not carefully designed to fuse deep features. The summary of all the above existing fusion methods is shown in Table \ref{tab:intro}.

In order to solve these drawbacks, we propose a novel fusion framework based on a novel connection architecture and an appropriate fusion strategy. The main contributions of our fusion framework are summarized as follows:

(1) The nest connection architecture \cite{zhou2018unet++} is applied to the CNN based fusion framework. Our nest connection-based framework is different from existing nest connection based framework. It contains three parts: encoder network, fusion strategy and decoder network respectively.

(2) Our nest connection architecture makes full use of deep features and preserves more information from different scale features which are extracted by the encoder network.

(3) For the fusion of multi-scale deep features, we propose a novel fusion strategy based on spatial attention and channel attention models. 

(4) Compared with existing state-of-the-art fusion methods, our fusion framework has better performance in terms of both visual assessment and objective assessment.

The rest of our paper is structured as follows. In Section \ref{related}, we briefly review related works on deep learning-based fusion methods. In Section \ref{proposed} we present the proposed fusion framework in detail. And in Section \ref{exp} we illustrate the experimental results. Finally, we draw conclusion in section \ref{con}.

%------------------------------------------------------------------------

\section{Related Works}
\label{related}

With the rise of deep learning in recent years, a lot of deep learning based methods have been proposed for the image fusion task. These methods attempt to design an end-to-end network to directly generate fused images \cite{liu2018deep}. In this section, firstly, we briefly introduce several classical methods, and the latest deep learning-based methods. Then we present the nest connection approach.

\subsection{Deep Learning-based Fusion Methods}

In 2017, a CNN-based fusion network was proposed by Liu et al. \cite{liu2017multi}. In their paper, the pairs of image patches ($16\times 16$) which contain different blur versions were used to train their network. The label of clear patch and blur patch were 1 and 0, respectively. The aim of this network was to generate a decision map, which indicated which source image is in more focus at the corresponding points. With the training phase, this CNN-based method has obtained better fusion performance than other algorithms before 2017. However, due to the limitation of training strategy, this method is only suitable for multi-focus images.

To overcome this weakness, Li et al. \cite{li2018densefuse} proposed a novel auto-encoder based network (DenseFuse) for fusing infrared and visible images. It consists of three parts: encoder, fusion layer and decoder. In the training phase, the fusion layer is discarded and the DenseFuse degenerates into an auto-encoder network. The purpose of the training phase was to obtain two sub-networks in which the encoder fully extracted deep features from source images and decoder adaptively reconstructed the raw data according to the encoded features. During the testing phase, the fusion layer was utilized to fuse deep features. Then, the fused image was reconstructed by the decoder network. To preserve more detail information, Zhang et al. \cite{zhang2020ifcnn} proposed a general end-to-end fusion network which was a simple yet effective architecture to generate fused images.

The GAN architecture was introduced to the infrared and visible image fusion field (FusionGAN) by Ma et al. \cite{ma2019fusiongan}. In the training phase, the source images were concatenated as a tensor to feed into the generator network, and the fused image was obtained by this network. Their loss function contained two terms: content loss and discriminator loss. With the adversarial strategy, the generator network can be trained to fuse arbitrary infrared and visible images.

%The content loss makes the fused image contain more infrared features and its gradient more like visible images. The discriminator loss is calculated by discriminator network which constrains the fused image to be more like the visible image.

%--------------------------------------------------------------------------------------------------
\subsection{The Nest Connection Architecture}

\begin{figure}[ht]
\centering
\includegraphics[width=0.8\linewidth]{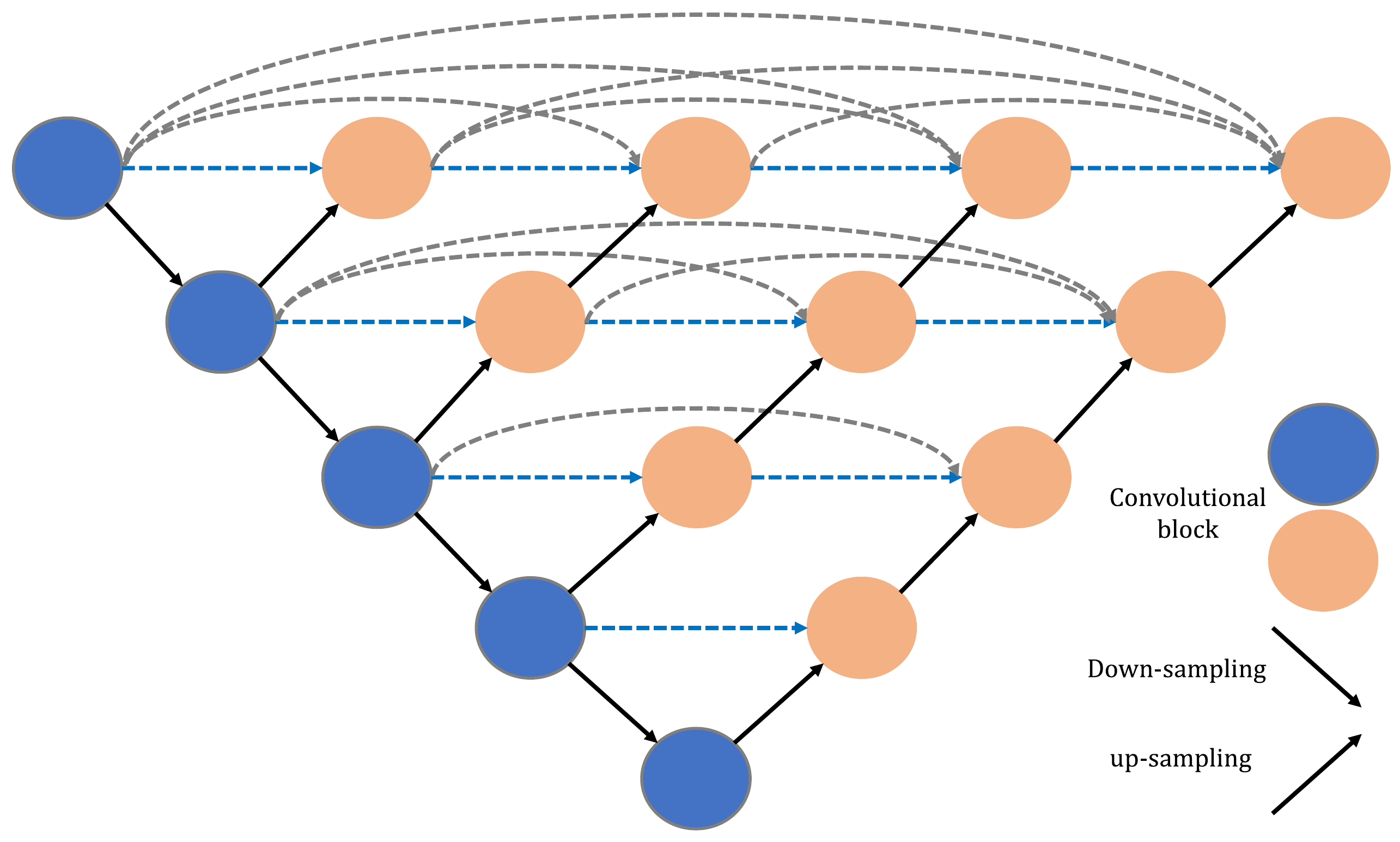}
\caption{The architecture of nest connection in UNet++.}
\label{fig:unet}
\end{figure}

The nest connection architecture was proposed by Zhou et al. \cite{zhou2018unet++} for the task of medical image segmentation. In the deep learning network, skip connection is a common operator to preserve more information from previous layers. However, the semantic gap causes unexpected results when long skip connections are used in the network architecture. To solve this problem, Zhou et al. presented a novel architecture (nest connection) which uses up-sampling and several short skip connections to replace a long skip connection. The framework of nest connection is illustrated in Fig.\ref{fig:unet}.

With the nest connection, the influence of the semantic gap is constrained and more information is preserved to obtain better segmentation results.

Inspired by this work, we introduce this architecture into the image fusion task and propose a modified nest connection based fusion framework based on nest connection and a novel fusion strategy.

%-------------------------------------------------------------------------
\section{Proposed Fusion Method}
\label{proposed}

In this section, the proposed nest connection-based fusion network is introduced in detail. Firstly, the fusion framework is presented in section \ref{fusion-network}. Then, the detail of training phase is described in section \ref{train-phase}. Finally, we present our novel fusion strategy based on two stages of attention models.

\begin{figure}[ht]
\centering
\includegraphics[width=\linewidth]{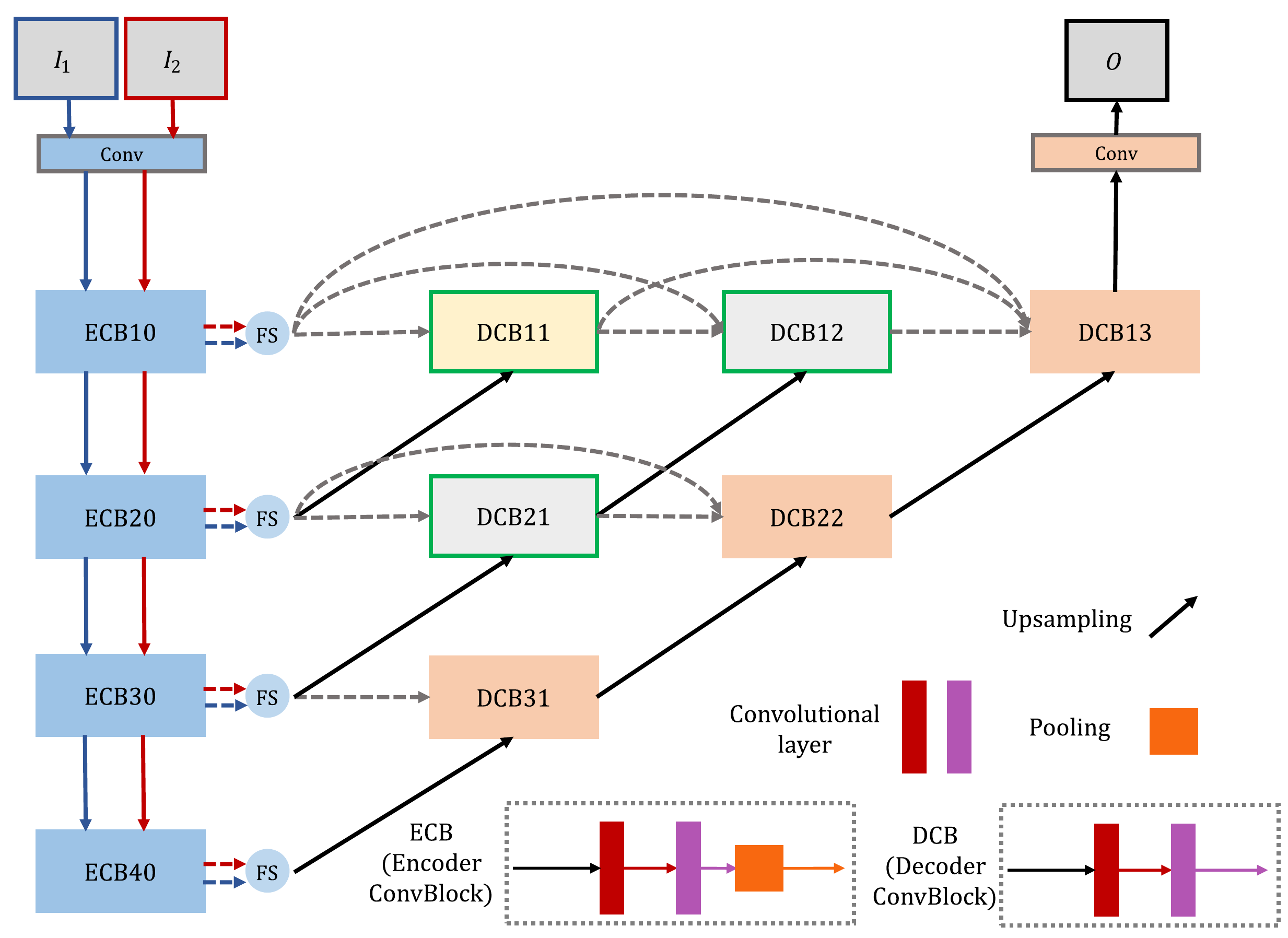}
\caption{The framework of proposed method. ``FS'' indicates fusion strategy.}
\label{fig:framework}
\end{figure}

%--------------------------------------------------------------------------------------------------
\subsection{Fusion Network}
\label{fusion-network}

Our fusion network (see Fig.\ref{fig:framework}) contains three main parts: encoder (blue square), fusion strategy (blue circle) and decoder (others), respectively. The nest connection is utilized in decoder network to process multi-scale deep features which are extracted by the encoder.

In Fig.\ref{fig:framework}, $I_1$ and $I_2$ indicate the source images. $O$ denotes the fused image. ``Conv'' means one convolutional layer. ``ECB'' denotes encoder convolutional block which contains two convolutional layers and one max-pooling layer. And ``DCB'' indicates decoder convolutional block without pooling operator.

Firstly, two input images are separately fed into encoder network to get multi-scale deep features. For each scale features, our fusion strategy is utilized to fuse the resulting features. Finally, the nest connection-based decoder network is used to reconstruct the fused image using the fused multi-scale deep features.

In next sections, we will introduce the training phase and the novel fusion strategy, respectively.

%--------------------------------------------------------------------------------------------------
\subsection{Training Phase}
\label{train-phase}

\begin{figure}[ht]
\centering
\includegraphics[width=\linewidth]{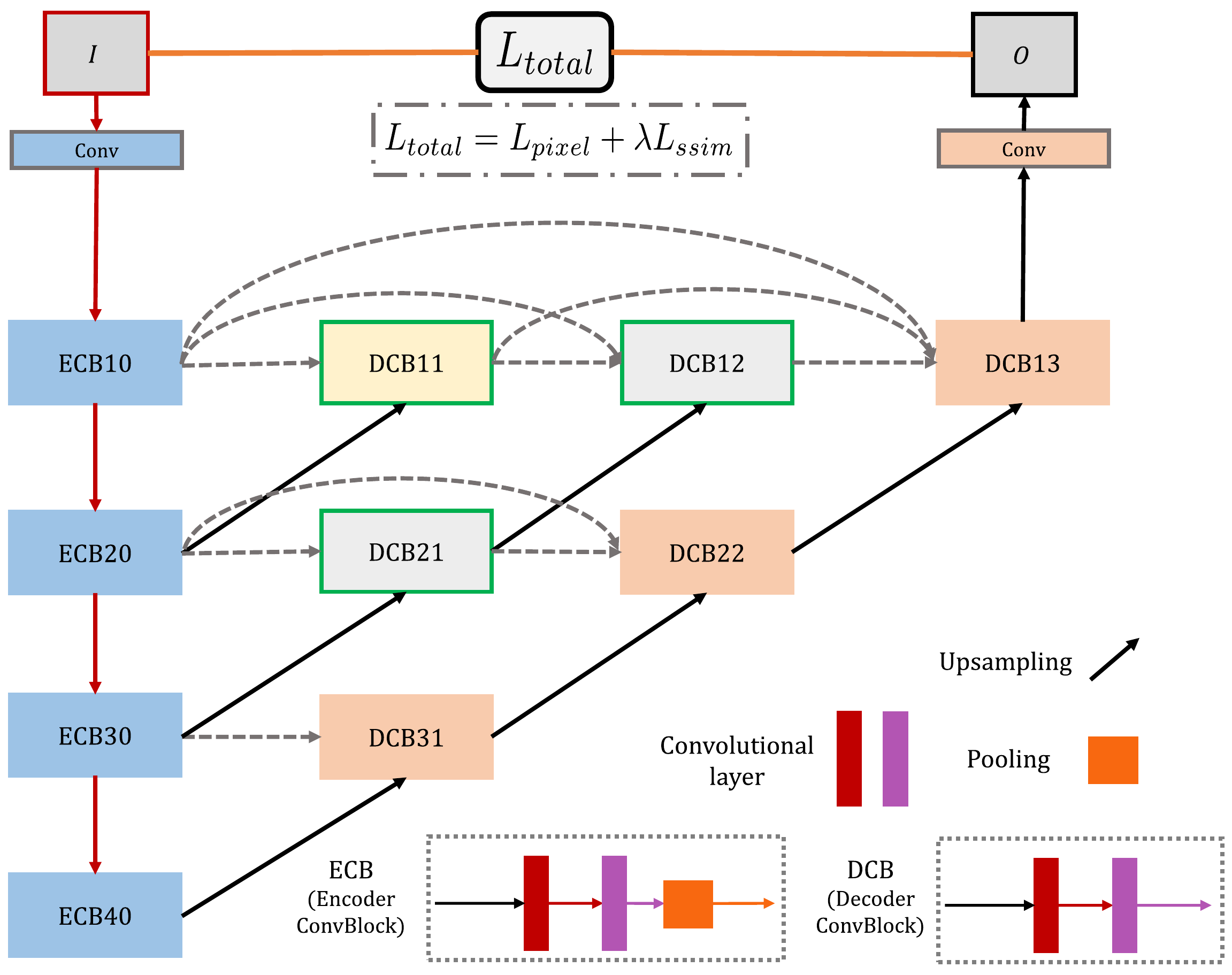}
\caption{The framework of training process.}
\label{fig:train}
\end{figure}

The training strategy is similar to the DenseFuse \cite{li2018densefuse}. In the training phase, the fusion strategy is discarded. We want to train an auto-encoder network in which the encoder is able to extract multi-scale deep features and the decoder reconstructs the input image from these features. The training framework is shown in Fig.\ref{fig:train}, and the fusion network settings are outlined in Table \ref{tab:trainframe}.

In Fig.\ref{fig:train} and Table \ref{tab:trainframe}, $I$ and $O$ are input image and output image, respectively. The encoder network consists of one convolutional layer (``Conv'') and four convolutional blocks (``ECB10'', ``ECB20'', ``ECB30'' and ``ECB40''). Each block contains two convolutional layers and one max-pooling operator which can ensure that encoder network can extract deep features in different scales.

The decoder network has six convolutional blocks (``DCB11'', ``DCB12'', ``DCB13''; ``DCB21'', ``DCB22''; ``DCB31'') and one convolutional layer (``Conv''). Six convolutional blocks are connected by nest connection architecture to avoid the semantic gap between encoder and decoder.

\begin{table}[!ht]
\centering
\caption{\label{tab:trainframe}The network settings of encoder and decoder network. \emph{\textbf{Conv}} is convolutional layer; \emph{\textbf{ECB}} denotes the encoder convolutional block (convolutional layer + max pooling); \emph{\textbf{DCB}} denotes decoder convolutional block (without pooling); the values of $N_{in}$ and $N_{out}$ depend on which layer of ``\emph{ECB}'' or ``\emph{DCB}'' belongs in encoder or decoder.}
\resizebox{0.9\linewidth}{!}{
\begin{tabular}{|c|c|c|c|c|c|c|}
\hline
 & Layer & Size & Stride & \makecell[cc]{Channel\\(input)} & \makecell[cc]{Channel\\(output)} & Activation \\
\hline
\multirow{5}*{Encoder} &
Conv          & 3      & 1      & 1      	& 16      	 & ReLu \\
~ & ECB10     & -      & -      & 16       	& 64         & - \\
~ & ECB20     & -      & -      & 64       	& 112        & - \\
~ & ECB30     & -      & -      & 112       & 160        & - \\
~ & ECB40     & -      & -      & 160       & 208        & - \\
\hline
\multirow{7}*{Decoder} &
    DCB31     & -      & -      & 368     & 160      & - \\
~ & DCB21     & -      & -      & 272     & 112      & - \\
~ & DCB22     & -      & -      & 384     & 112      & - \\
~ & DCB11     & -      & -      & 176     & 64       & - \\
~ & DCB12     & -      & -      & 240     & 64       & - \\
~ & DCB13     & -      & -      & 304     & 64       & - \\
~ & Conv      & 1      & 1      & 64      & 1        & ReLu \\
\hline
\multirow{3}*{ECB} &
    Conv     	& 3      & 1      & $N_{in}$     & 16       & ReLu \\
~ & Conv     	& 3      & 1      & 16     & $N_{out}$      & ReLu \\
~ & max-pooling & -      & -      & -      & -       		& -\\
\hline
\multirow{2}*{DCB} &
    Conv     	& 3      & 1      & $N_{in}$     & 16       & ReLu \\
~ & Conv     	& 3      & 1      & 16     & $N_{out}$      & ReLu \\
\hline
\end{tabular}}
\end{table}

In training phase, the loss function $L_{total}$ is defined as follows,
\begin{eqnarray}\label{Eq1}
  	L_{total} = L_{pixel} + \lambda L_{ssim}
\end{eqnarray}
\noindent where $L_{pixel}$ and $L_{ssim}$ indicate the pixel loss and structure similarity ($SSIM$) loss between the input image $I$ and the output image $O$. $\lambda$ denotes the trade-off value between $L_{pixel}$ and $L_{ssim}$.

$L_{pixel}$ is calculated by Eq.\ref{Eq2},
\begin{eqnarray}\label{Eq2}
  	L_{pixel} = ||O-I||_F^2
\end{eqnarray}
\noindent where $O$ and $I$ indicate the output and input images, respectively. $||\cdot||_F$ is the Frobenius norm. $L_{pixel}$ calculates the distance between $O$ and $I$. This loss function will make sure that the reconstructed image is more similar to input image in pixel level.

The SSIM loss $L_{ssim}$ is obtained by Eq.\ref{Eq3},
\begin{eqnarray}\label{Eq3}
  	L_{ssim} = 1-SSIM(O,I)
\end{eqnarray}
\noindent where $SSIM(\cdot)$ denotes the structural similarity measure \cite{wang2004image}. The output image $O$ and the input image $I$ have more similarity in structure when the values of $SSIM(\cdot)$ become larger. 

The aim of the training phase is to obtain two powerful tools for the encoder network and the decoder network. Thus, the type of input images in training phase is not limited to infrared and visible images. In the training stage, the dataset MS-COCO \cite{lin2014microsoft} is used to train our auto-encoder network and we choose 80000 images to be the input images. These images are converted to gray scale and then resized to $256\times 256$. As the orders of magnitude are different between $L_{pixel}$ and $L_{ssim}$, the parameter $\lambda$ is set as 1, 10, 100 and 1000 to train our network. The detailed analysis of training phase is introduced in the Ablation Study given in Section \ref{abstudy}.

%--------------------------------------------------------------------------------------------------
\subsection{Fusion Strategy}
\label{FS}

Most fusion strategies are based on the weight-average operator which generates a weighting map to fuse the source images. Based on this theory, the choice of the weighting map becomes a key issue.

The fusion network becomes more flexible when the fusion strategies are added to the test phase \cite{li2018densefuse}, however, these strategies are not designed for deep features, and attention mechanism is not considered yet.

\begin{figure}[ht]
\centering
\includegraphics[width=0.9\linewidth]{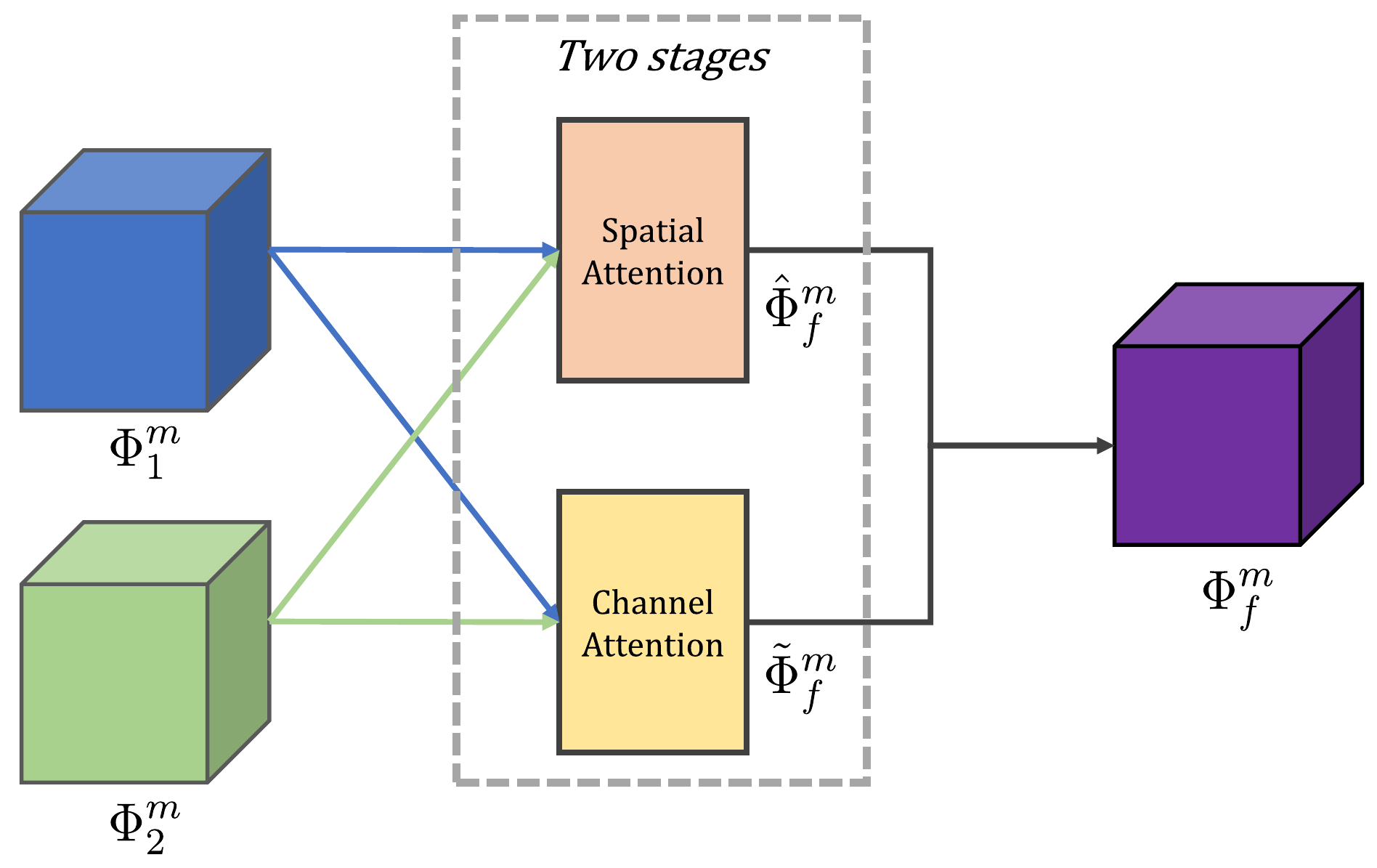}
\caption{The procedure of attention model-based fusion strategy.}
\label{fig:fs}
\end{figure}

To solve this problem, in this section, we introduce a novel fusion strategy based on two stages of attention models. In our fusion architecture, $m$ indicates the level of multi-scale deep features and $m\in{\{1,2,\cdots,M\}},M=4$. The framework of our fusion strategy is shown in Fig.\ref{fig:fs}.

$\Phi_1^m$ and $\Phi_2^m$ are multi-scale deep features which are extracted by encoder from two input images, respectively. $\hat{\Phi}_f^m$ and $\tilde{\Phi}_f^m$ are fused features which are obtained by spatial attention model and channel attention model, respectively. $\Phi_f^m$ is the final fused multi-scale deep feature which will be the input of the decoder network.

In our fusion strategy, we focus on two types of features: spatial attention model and channel attention model. The extracted multi-scale deep features are processed in two phases.

When $\hat{\Phi}_f^m$ and $\tilde{\Phi}_f^m$ are obtained by our attention models, the final features are generated by Eq.\ref{Eq4},
\begin{eqnarray}\label{Eq4}
  	\Phi_f^m = (\hat{\Phi}_f^m + \tilde{\Phi}_f^m) \times 0.5
\end{eqnarray}

Now, we will introduce our attention model-based fusion strategies in detail.

\subsubsection{\textbf{Spatial Attention Model}}
\label{FS-spa}

\begin{figure}[ht]
\centering
\includegraphics[width=\linewidth]{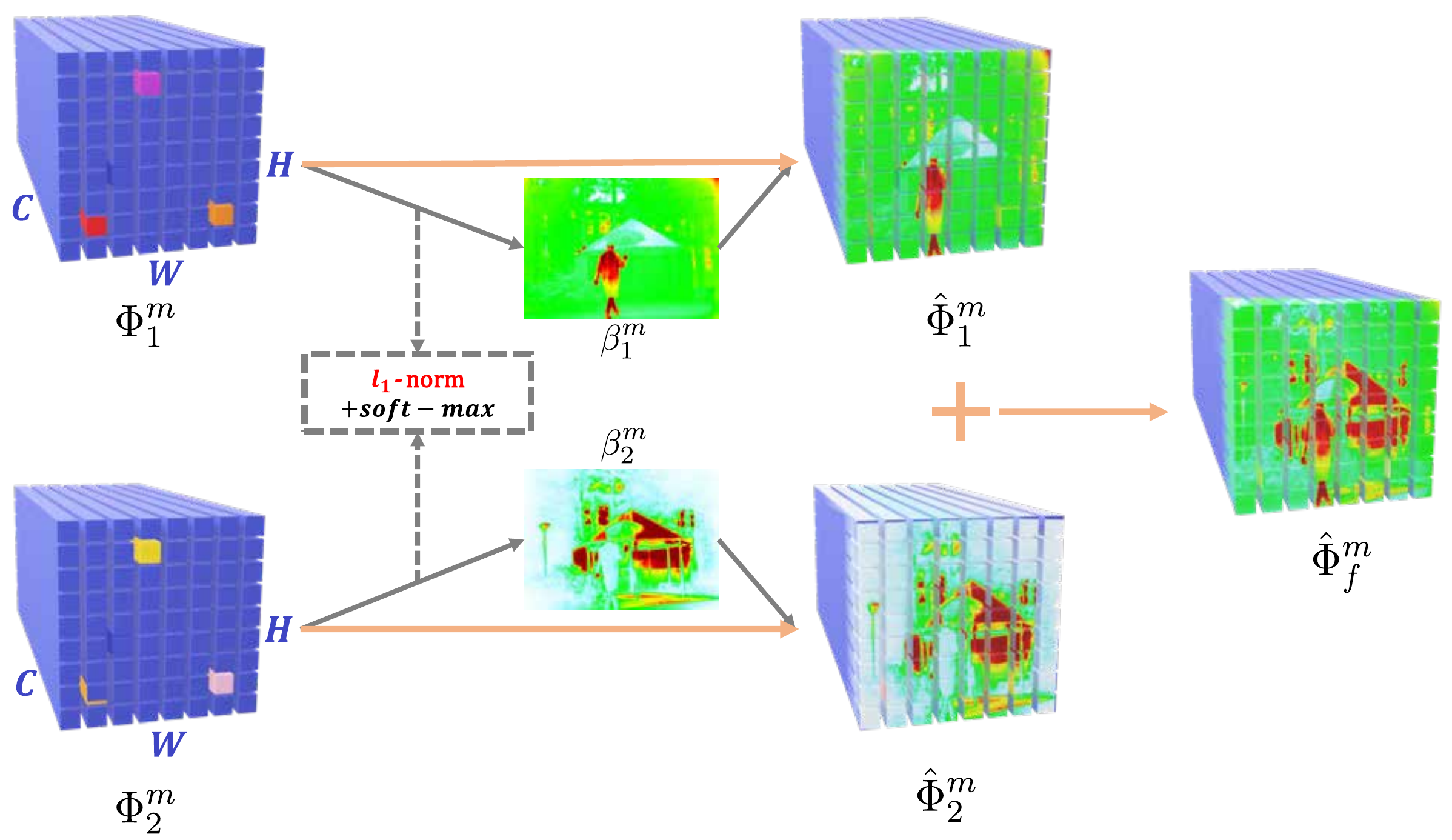}
\caption{The procedure of spatial attention-based fusion strategy.}
\label{fig:spatial}
\end{figure}

In \cite{liu2016image}\cite{li2018infrared}\cite{li2018densefuse}, a spatial-based fusion strategy is utilized in the image fusion task. In this paper, we extend this operation to fuse multi-scale deep features and is called the spatial attention model. The procedure for obtaining the spatial attention model is shown in Fig.\ref{fig:spatial}.

$\beta_1^m$ and $\beta_2^m$ indicate the weighting maps which are calculated by $l_1$-norm and soft-max operator from deep features $\Phi_1^m$ and $\Phi_2^m$. The weighting maps are formulated by Eq.\ref{Eq5},
\begin{eqnarray}\label{Eq5}
  	\beta_k^m(x,y) = \frac{||\Phi_k^m(x,y)||_1}{\sum_{i=1}^K||\Phi_i^m(x,y)||_1}
\end{eqnarray}

\noindent where $||\cdot||_1$ denotes $l_1$-norm, $k\in{1,\cdots,K}$ and $K=2$. $(x,y)$ indicates the corresponding position in multi-scale deep features ($\Phi_1^m$ and $\Phi_2^m$) and weighting maps ($\beta_1^m$ and $\beta_2^m$), each position denotes a $C$ dimensional vector in deep features. The $\Phi_k^m(x,y)$ denotes a vector which has $C$ dimensions.

$\hat{\Phi}_1^m$ and $\hat{\Phi}_2^m$ denote the enhanced deep features which are weighted by $\beta_1^m$ and $\beta_2^m$. The enhanced features($\hat{\Phi}_k^m$) are calculated by Eq.\ref{Eq6},	
\begin{eqnarray}\label{Eq6}
  	\hat{\Phi}_k^m(x,y) = \beta_k^m(x,y) \times \Phi_k^m(x,y)
\end{eqnarray}

Then the fused features $\hat{\Phi}_f^m$ are calculated by adding these enhanced deep features, the formulation is shown in Eq.\ref{Eq7},
\begin{eqnarray}\label{Eq7}
  	\hat{\Phi}_f^m(x,y) = \sum_{i=1}^K \hat{\Phi}_i^m(x,y)
\end{eqnarray}

\subsubsection{\textbf{Channel Attention Model}}

\begin{figure}[ht]
\centering
\includegraphics[width=\linewidth]{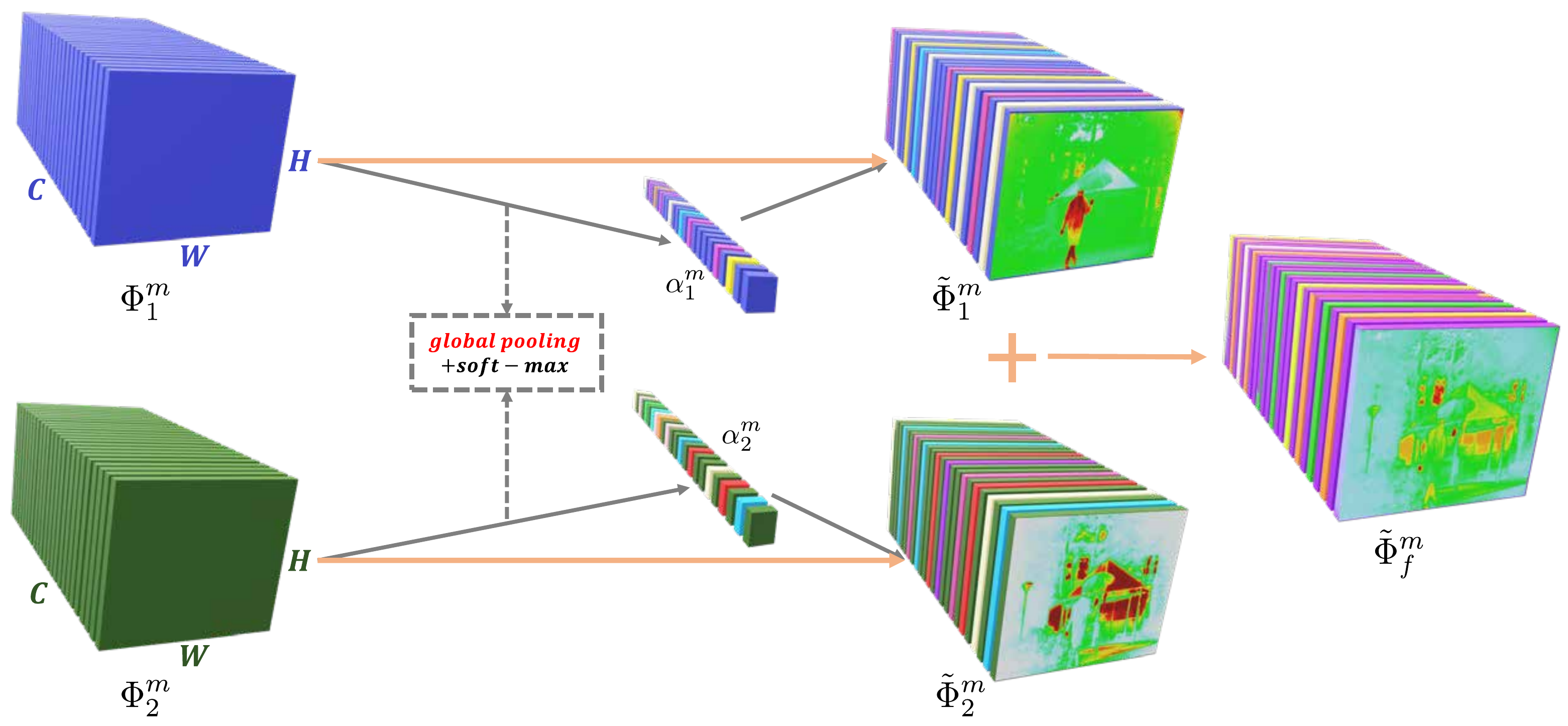}
\caption{The diagram of channel attention-based fusion strategy.}
\label{fig:channel}
\end{figure}

In existing deep learning-based fusion methods, almost fusion strategies just calculate the spatial information. However deep features are three dimensional tensors. Hence not only spatial dimensional information, but the channel information should also be considered in the fusion strategy as well. Thus, we propose a channel attention-based fusion strategy. The diagram of this strategy is shown in Fig.\ref{fig:channel}.

As we discussed in section \ref{FS-spa}, $\Phi_1^m$ and $\Phi_2^m$ are multi-scale deep features. $\alpha_1^m$ and $\alpha_2^m$ are C dimensional weighting vectors which are calculated by global pooling and soft-max. $\tilde{\Phi}_1^m$ and $\tilde{\Phi}_2^m$ indicate enhanced deep features which are weighted by weighting vectors. $\tilde{\Phi}_f^m$ is fused features which are calculated by channel attention-based fusion strategy.

Firstly, a global pooling operator is utilized to calculate the initial weighting vectors ($\bar{\alpha}_1^m$ and $\bar{\alpha}_2^m$). The formulation is shown in Eq.\ref{Eq8},
\begin{eqnarray}\label{Eq8}
  	\bar{\alpha}_k^m(n) = P(\Phi_k^m(n))
\end{eqnarray}
\noindent where $k\in{\{1,2\}}$, $n$ indicates the corresponding index of channel in deep features $\Phi_k^m$, $P(\cdot)$ is the global pooling operator. 

In our channel attention model, three global pooling operations are chosen, including: (1) Average operator which calculates the average values of each channel; (2) Max operator which calculates the maximum value of each channel; (3) Nuclear-norm operator ($||\cdot||_*$) which is the sum of singular values for one channel. The influence of different operators for global pooling will be discussed in the Ablation Study \ref{abstudy}.

Then, a soft-max operator (Eq.\ref{Eq9}) is used to obtain the final weighting vectors $\alpha_1^m$ and $\alpha_2^m$,
\begin{eqnarray}\label{Eq9}
  	\alpha_k^m(n) = \frac{\bar{\alpha}_k^m(n)}{\sum_{i=1}^K\bar{\alpha}_i^m(n)}
\end{eqnarray}

When we obtain the final weight vectors, the fused features which are generated by channel attention model can be calculated by Eq.\ref{Eq10},
\begin{eqnarray}\label{Eq10}
	\tilde{\Phi}_f^m(n) =& \sum_{i=1}^K \alpha_i^m(n)\times \Phi_i^m(n)
\end{eqnarray}

%------------------------------------------------------------------------
\section{Experimental Results}
\label{exp}

In this section, we first describe the experimental settings of testing phase. Then, we introduce our ablation study. We compare our method with other existing methods in subjective evaluation and utilize several quality metrics to evaluate the fusion performance objectively.

%The line charts of loss values are shown in Fig.\ref{fig:loss}.

\begin{figure}[ht]
\centering
\includegraphics[width=\linewidth]{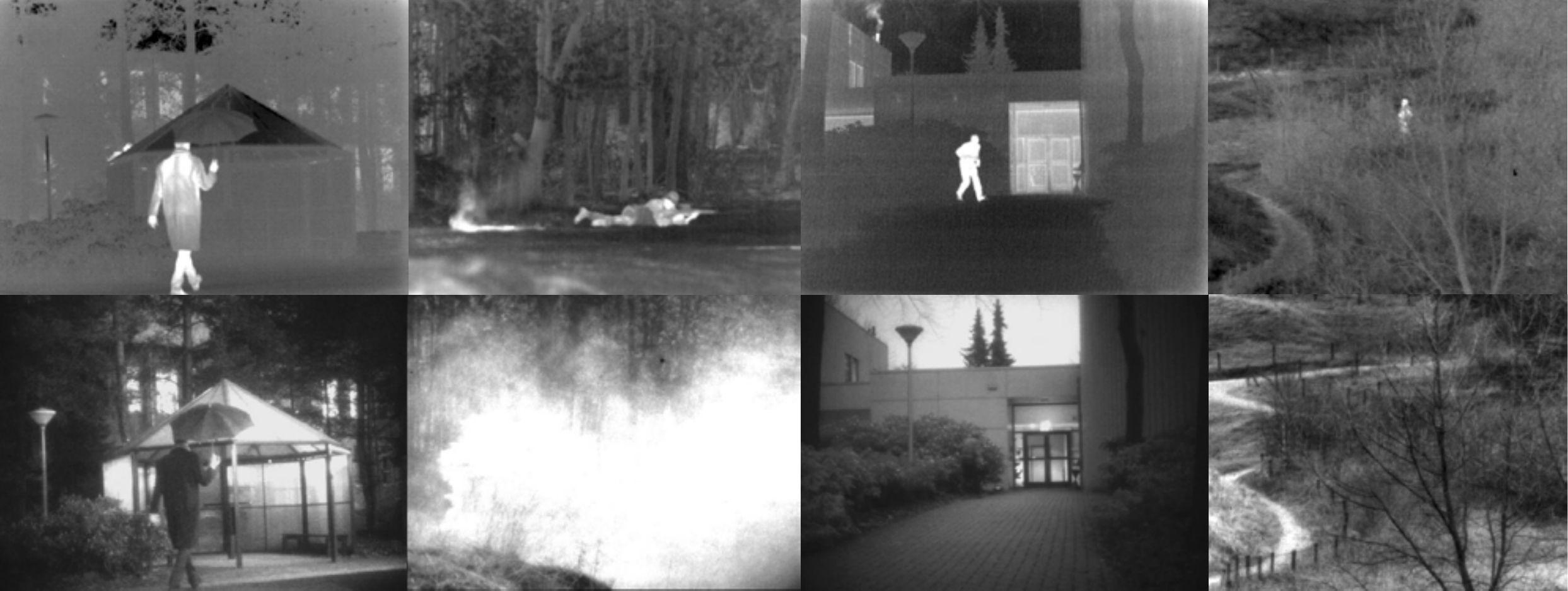}
\caption{Four pairs of source images. The top row contains infrared images, and the second row contains visible images.}
\label{fig:examples}
\end{figure}

%--------------------------------------------------------------------------------------------------
\subsection{Experimental Settings}

In our experiments, 21 pairs of infrared and visible images\footnote{These images are available at \url{https://github.com/hli1221/imagefusion-nestfuse}.} were collected from \cite{ma2017infrared} and \cite{tno2018}. A sample of these images is shown in Fig.\ref{fig:examples}.

We choose twelve typical and state-of-the-art fusion methods to evaluate the fusion performance, including: cross bilateral filter fusion method (CBF) \cite{kumar2015image}, discrete cosine harmonic wavelet transform fusion method (DCHWT) \cite{kumar2013multifocus}, joint SR based fusion method (JSR) \cite{zhang2013dictionary}, the joint sparse representation model with saliency detection fusion method (JSRSD) \cite{liu2017infrared}, gradient transfer and total variation minimization (GTF) \cite{ma2016infrared}, visual saliency map and weighted least square optimization based fusion method (WLS) \cite{ma2017infrared}, convolutional sparse representation based fusion method (ConvSR) \cite{liu2016image}, VGG-19 and the multi-layer fusion strategy-based method (VggML) \cite{li2018infrared}, DeepFuse \cite{ram2017deepfuse}, DenseFuse \cite{li2018densefuse}\footnote{The addition fusion strategy is utilized and the parameter $\lambda$ is set to 100.}, the GAN-based fusion network (FusionGAN) \cite{ma2019fusiongan} and a general end-to-end fusion network(IFCNN) \cite{zhang2020ifcnn}. All these comparison fusion methods are implemented based on their publicly available codes, and their parameters are set by referring to their papers.

Seven quality metrics are utilized for quantitative comparison between our fusion method and other existing fusion methods. These are: entropy ($En$) \cite{roberts2008assessment}; standard deviation ($SD$) \cite{rao1997fibre}; mutual information ($MI$) \cite{peng2005feature}; $FMI_{dct}$ and $FMI_w$ \cite{haghighat2014fast} which calculates mutual information ($FMI$) for the discrete cosine transform and the region feature; the modified structural similarity for no-reference image ($SSIM_a$); and visual information fidelity ($VIF$) \cite{han2013new}.

The $SSIM_a$ is calculated by Eq.\ref{Eq11},
\begin{eqnarray}\label{Eq11}
  	SSIM_a(F) = (SSIM(F,I_1)+SSIM(F,I_2))\times0.5
\end{eqnarray}
\noindent where $SSIM(\cdot)$ denotes the structural similarity measure \cite{wang2004image}, $F$ is fused image, and $I_1$, $I_2$ are source images.

The fusion performance improves with the increasing numerical index of all these seven metrics. The larger $En$ and $SD$ means input image contains more information, which also indicates that the fusion method achieves better performance. The larger $MI$, $FMI_{dct}$ and $FMI_{w}$ indicates the fusion method could preserve more raw information and features from source images. For $SSIM_a$ and $VIF$, the fusion algorithms preserve more structural information from source images and generate more natural features.

%--------------------------------------------------------------------------------------------------
\subsection{Ablation Study}
\label{abstudy}

\subsubsection{\textbf{Parameter($\lambda$) in Loss Function}}

\begin{figure*}[!ht]
\centering
\includegraphics[width=0.9\linewidth]{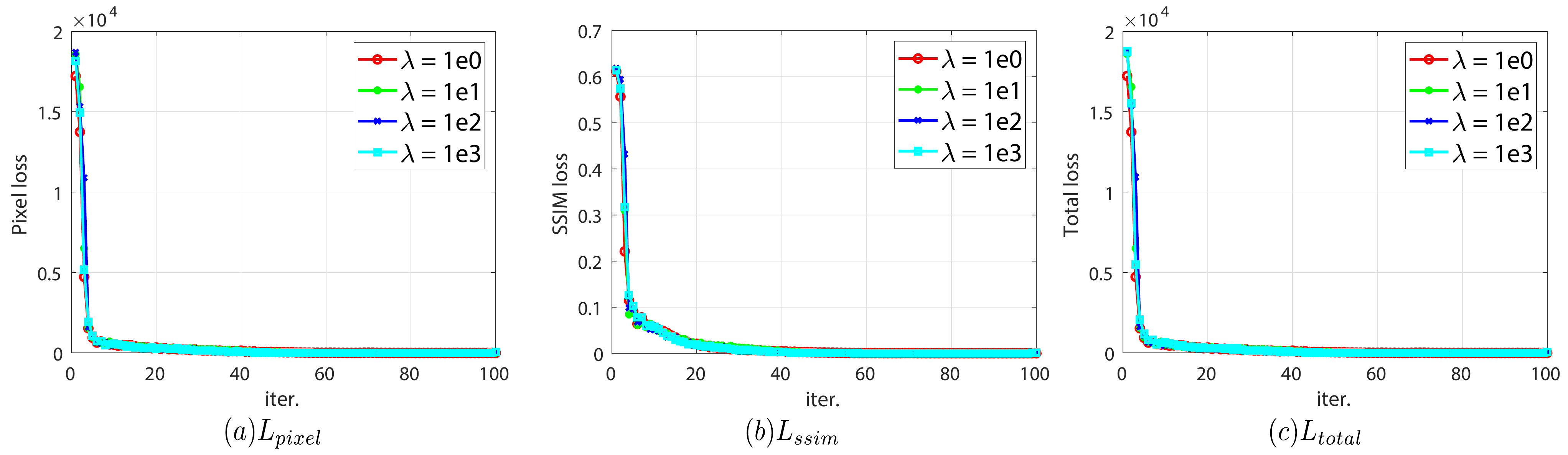}
\caption{The line charts of pixel loss(a), SSIM loss(b) and total loss(c) in training phase. Each point in horizontal axis indicates 10 iterations and we choose the first 1000 iterations.}
\label{fig:loss}
\end{figure*}

As discussed in section \ref{train-phase}, the parameter $\lambda$ is set as 1, 10, 100 and 1000. The epoch and batch size are 2 and 4, respectively. Our network is implemented with NVIDIA GTX 1080Ti and PyTorch is used for implementation. The line chart of loss values is demonstrated in Fig.\ref{fig:loss}.

In Fig.\ref{fig:loss}, at first 400 iterations, the auto-encoder network has rapid convergence with the increase of the parameter $\lambda$. $L_{pixel}$, $L_{ssim}$ and $L_{total}$ have faster convergence rate when $\lambda=100$ or $\lambda=1000$. In addition, when iterations are more than 600, we get the optimal network weights, no matter which $\lambda$ is chosen. In general, our fusion network gets faster convergence of $L_{ssim}$ with increase of $\lambda$ in the early stage.

We still need to choose one $\lambda$ values for our image fusion task based on the test images. Seven metrics are used to evaluate the performance of different network with different $\lambda$. And the operations $P(\cdot)$ which were utilized in channel attention model are $avg$, $max$ and $nuclear -norm$, respectively. These values are shown in Table \ref{tab:lambda}. The best values are indicated in \textbf{bold} and the second-best values are denoted in \emph{\color{red}{red and italic}}.

From Table \ref{tab:lambda}, although different $\lambda$ has no effect for convergence rate when iterations become larger, it still has influence for the fusion performance of our fusion framework. When $\lambda$ is 100 ($1e2$), our network can achieve better fusion performance than other values of $\lambda$. So, in our experiment, $\lambda$ is set as 100.

\begin{figure}[!ht]
\centering
\includegraphics[width=0.7\linewidth]{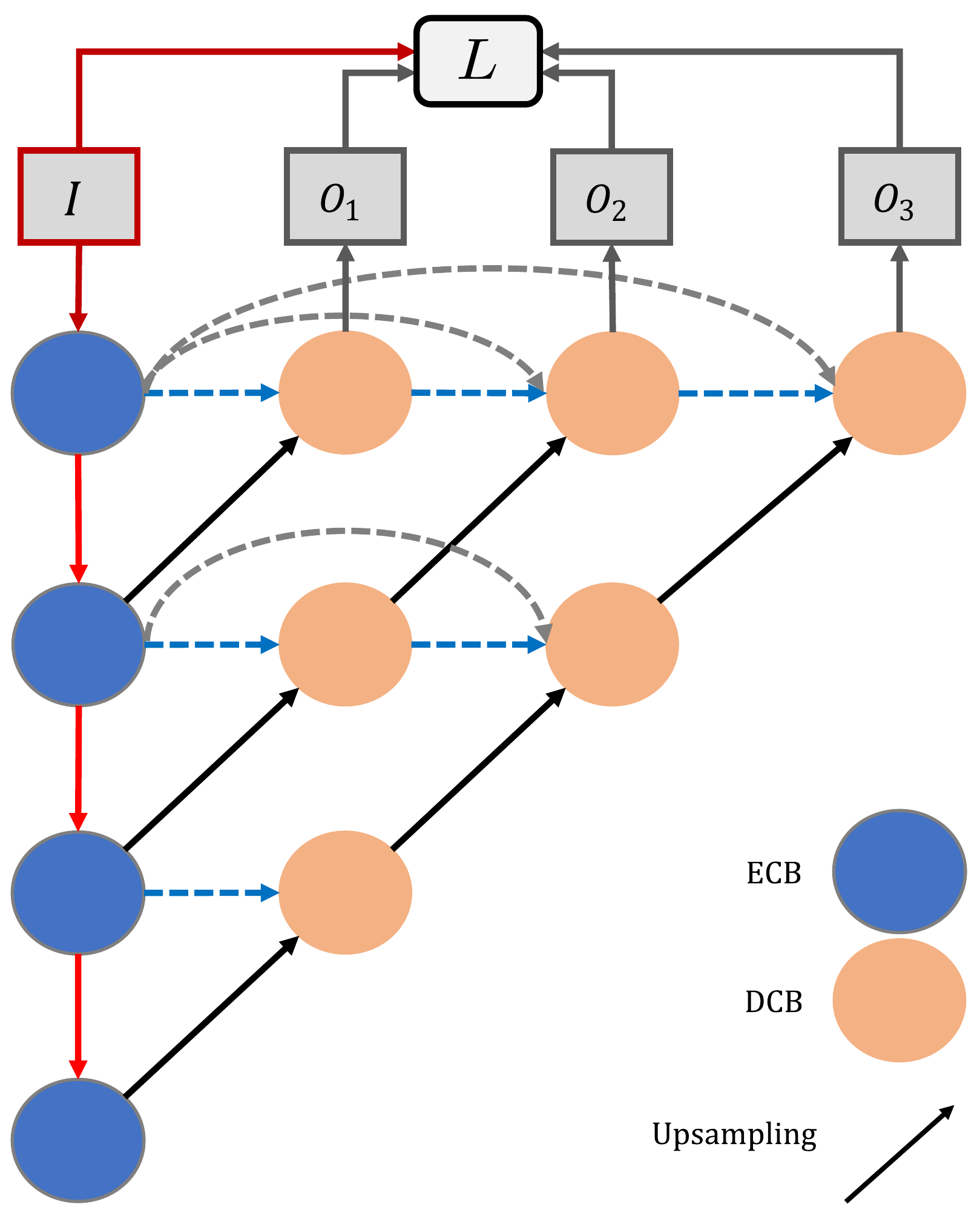}
\caption{The training framework of deep supervision. $O_1$, $O_2$ and $O_3$ are outputs based on different scale features.}
\label{fig:deeply-super}
\end{figure}

\begin{table*}[!ht]
\centering
\caption{\label{tab:lambda}The metrics values with different $\lambda$ and different global operations.}
\resizebox{0.9\linewidth}{!}{
\begin{tabular}{|c|c|c|c|c|c|c|c|c|c|}
\hline
 & $\lambda$ & $P(\cdot)$ & $En$\cite{roberts2008assessment} & $SD$\cite{rao1997fibre} & $MI$\cite{peng2005feature} & $FMI_dct$\cite{haghighat2014fast} & $FMI_w$\cite{haghighat2014fast} & $SSIM_a$ & $VIF$\cite{han2013new}\\
\hline
\multirow{12}*{NestFuse} &
\multirow{3}*{$1e0$} &$avg$		&\emph{\color{red}{6.91369}} 	&82.39563 	&\emph{\color{red}{13.82738}} 	&0.35118 	&0.43607 	&0.73154 	&\emph{\color{red}{0.78185}} \\
~	&~			     &$max$		&6.88793 	&80.02630 	&13.77586 	&0.35450 	&0.43172 	&0.73512 	&0.74792 \\
~	&~			     &$nuclear$	&6.89778 	&82.51583 	&13.79557 	&0.35700 	&0.43522 	&0.73323 	&0.75936 \\
\cline{2-10}
~& \multirow{3}*{$1e1$} &$avg$		&6.90552 	&81.79393 	&13.81103 	&0.34487 	&0.43387 	&0.73177 	&0.77405 \\
~	&~			     &$max$		&6.88281 	&79.80942 	&13.76562 	&0.34749 	&0.42985 	&0.73516 	&0.74333 \\
~	&~			     &$nuclear$	&6.89021 	&82.11951 	&13.78042 	&0.34967 	&0.43307 	&0.73339 	&0.75367 \\
\cline{2-10}
~& \multirow{3}*{$1e2$} &$avg$		&\textbf{6.91971} 	&\textbf{82.75242} 	&\textbf{13.83942} 	&\emph{\color{red}{0.35801}} 	&\textbf{0.43724} 	&0.73199 	&\textbf{0.78652} \\
~	&~			     &$max$		&6.89421 	&80.36372 	&13.78842 	&0.36080 	&0.43293 	&\emph{\color{red}{0.73532}} 	&0.75204 \\
~	&~			     &$nuclear$	&6.90461 	&\emph{\color{red}{82.92572} }	&13.80923 	&\textbf{0.36277} 	&\emph{\color{red}{0.43621}} 	&0.73360 	&0.76415 \\
\cline{2-10}
~& \multirow{3}*{$1e3$} &$avg$		&6.91062 	&82.14301 	&13.82125 	&0.34819 	&0.43572 	&0.73211 	&0.77881 \\
~	&~			     &$max$		&6.88939 	&80.18043 	&13.77877 	&0.35079 	&0.43109 	&\textbf{0.73547}	&0.74746 \\
~	&~			     &$nuclear$	&6.89612 	&82.40198 	&13.79224 	&0.35295 	&0.43437 	&0.73393 	&0.75632 \\
\hline
\end{tabular}}
\end{table*}

\begin{table*}[!ht]
\centering
\caption{\label{tab:deep-super} The objective evaluation of the outputs of deep supervision and without(w/o) deeply supervision.}
\resizebox{0.9\linewidth}{!}{
\begin{tabular}{|c|c|c|c|c|c|c|c|c|c|}
\hline
\multicolumn{3}{|c|}{} & $En$\cite{roberts2008assessment} & $SD$\cite{rao1997fibre} & $MI$\cite{peng2005feature} & $FMI_{dct}$\cite{haghighat2014fast} & $FMI_w$\cite{haghighat2014fast} & $SSIM_a$ & $VIF$\cite{han2013new}\\
\hline
\multirow{9}*{{\shortstack{deep\\ supersion}}} &
\multirow{3}*{$O_1$} &
  $avg$		&6.90495 	&81.95101 	&13.80989 	&0.34559 	&0.43414 	&0.73149 	&0.77568\\ 
&&$max$		&6.88467 	&80.00623 	&13.76934 	&0.34870 	&0.43018 	&0.73479 	&0.74551\\ 
&&$nuclear$	&6.89084 	&82.19268 	&13.78168 	&0.35087 	&0.43337 	&0.73322 	&0.75491\\ 
\cline{2-10}
&\multirow{3}*{$O_2$} &
  $avg$		&6.91023 	&82.31554 	&13.82046 	&0.34433 	&0.43395 	&0.73182 	&0.77754\\ 
&&$max$		&6.88734 	&80.10722 	&13.77469 	&0.34725 	&0.42984 	&0.73520 	&0.74612\\ 
&&$nuclear$	&6.89418 	&82.53775 	&13.78835 	&0.34941 	&0.43318 	&0.73350 	&0.75650\\
\cline{2-10}
&\multirow{3}*{$O_3$} &
  $avg$		&6.90866 	&82.23202 	&13.81732 	&0.34462 	&0.43399 	&0.73172 	&0.77702\\ 
&&$max$		&6.88544 	&80.00264 	&13.77088 	&0.34753 	&0.42984 	&0.73512 	&0.74526\\ 
&&$nuclear$	&6.89351 	&82.47415 	&13.78702 	&0.34965 	&0.43307 	&0.73334 	&0.75639\\ 
\hline
\multicolumn{2}{|c|}{\multirow{3}*{{\shortstack{w/o\\ deep supersion}}}} &
						$avg$		&\textbf{6.91971} 	&82.75242 	&\textbf{13.83942} 	&0.35801 	&\textbf{0.43724} 	&0.73199 	&\textbf{0.78652}\\ 
\multicolumn{2}{|c|}{} &$max$		&6.89421 	&80.36372 	&13.78842 	&0.36080 	&0.43293 	&\textbf{0.73532} 	&0.75204\\ 
\multicolumn{2}{|c|}{} &$nuclear$	&6.90461 	&\textbf{82.92572} 	&13.80923 	&\textbf{0.36277} 	&0.43621 	&0.73360 	&0.76415\\ 
\hline
\end{tabular}}
\end{table*}

\begin{figure*}[!ht]
\centering
\includegraphics[width=0.95\linewidth]{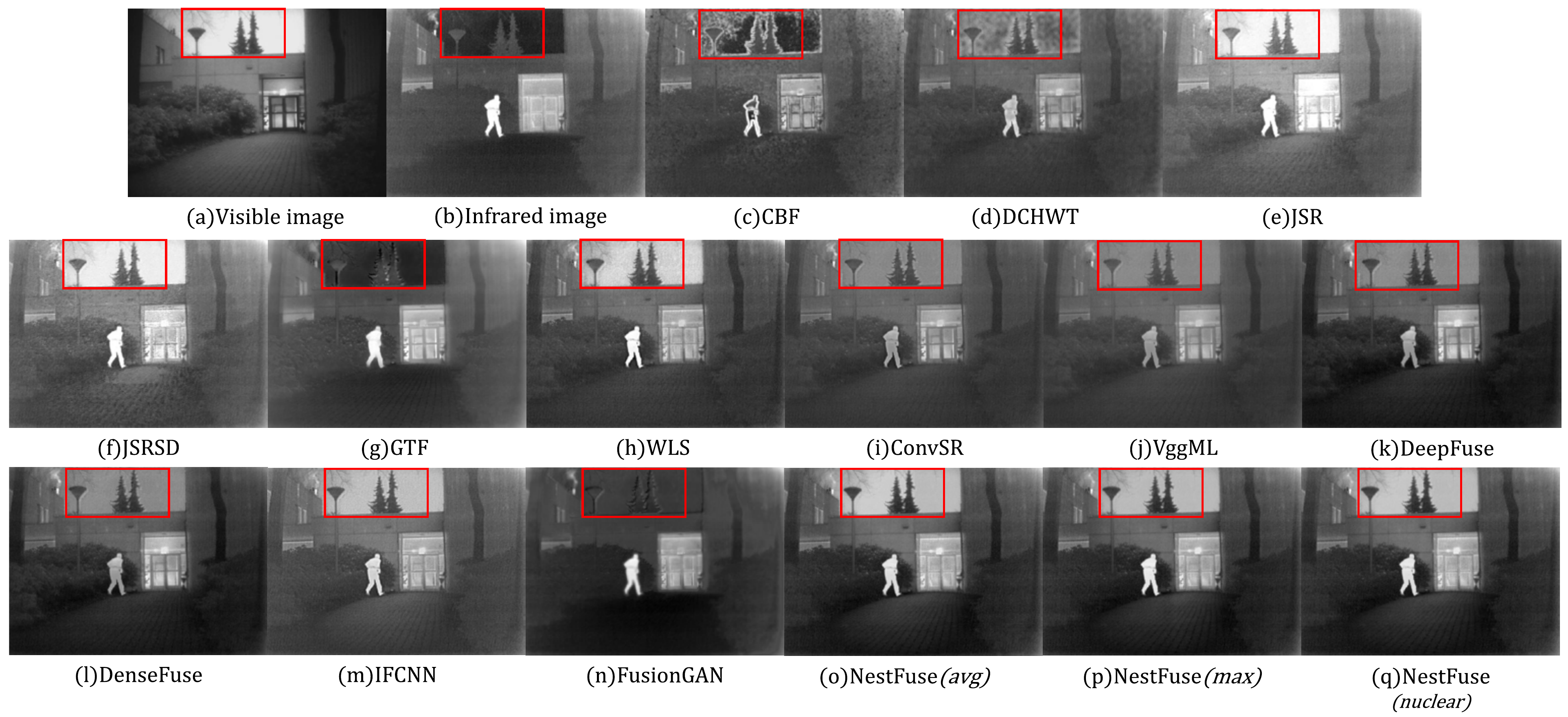}
\caption{Experiment on “man” images.  (a) Infrared image; (b) Visible image; (c) CBF; (d) DCHWT; (e) JSR; (f) JSRSD; (g) GTF; (h) WLS; (i) ConvSR; (j) VggML; (k) DeepFuse; (l) DenseFuse; (m) IFCNN; (n) FusionGAN; (o) NestFuse$(avg)$; (p) NestFuse$(max)$; (q) NestFuse$(nuclear)$.}
\label{fig:man}
\end{figure*}

\begin{figure*}[!ht]
\centering
\includegraphics[width=0.95\linewidth]{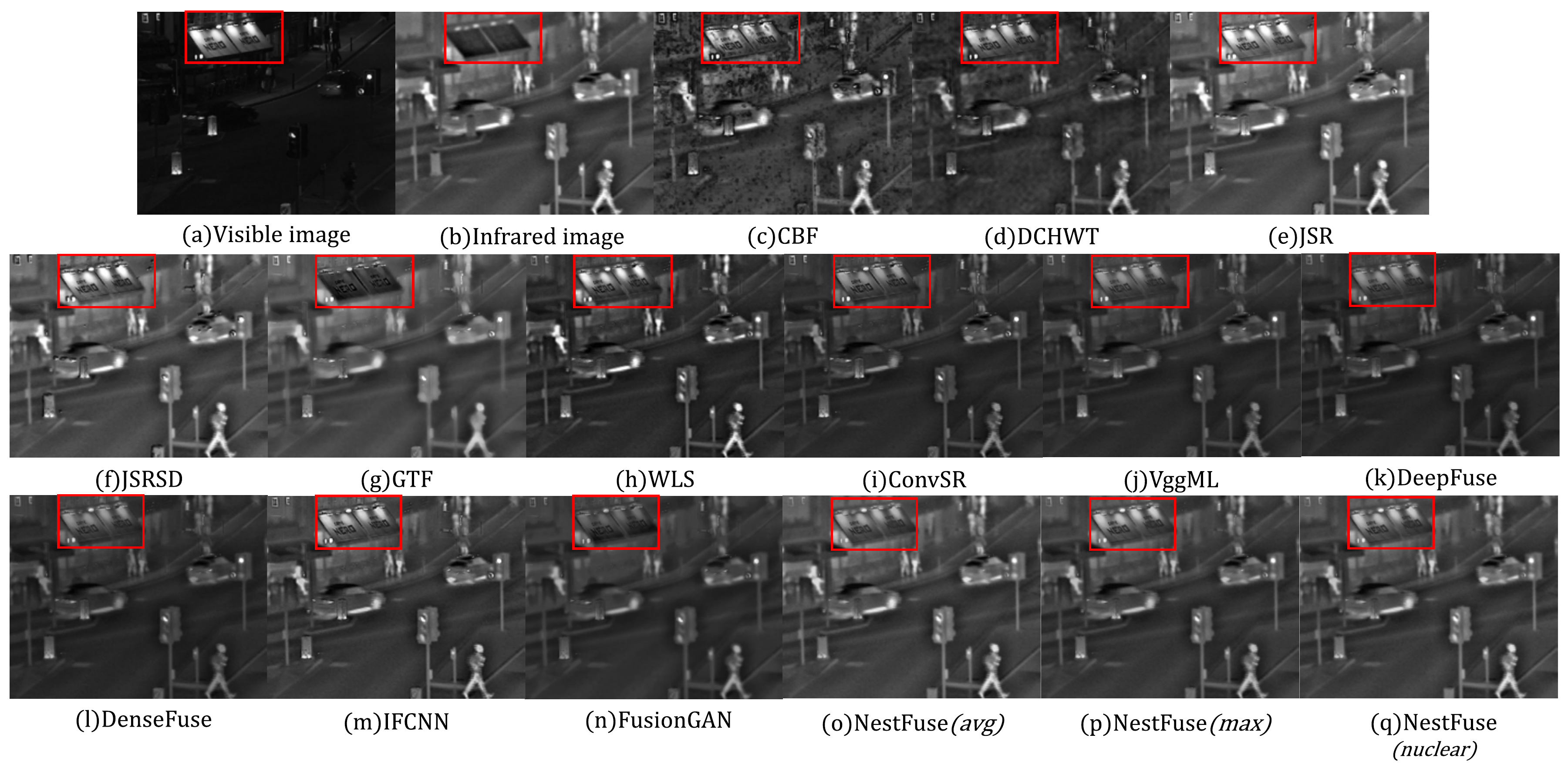}
\caption{Experiment on “street” images. (a) Infrared image; (b) Visible image; (c) CBF; (d) DCHWT; (e) JSR; (f) JSRSD; (g) GTF; (h) WLS; (i) ConvSR; (j) VggML; (k) DeepFuse; (l) DenseFuse; (m) IFCNN; (n) FusionGAN; (o) NestFuse$(avg)$; (p) NestFuse$(max)$; (q) NestFuse$(nuclear)$.}
\label{fig:street}
\end{figure*}

\begin{figure*}[!ht]
\centering
\includegraphics[width=0.9\linewidth]{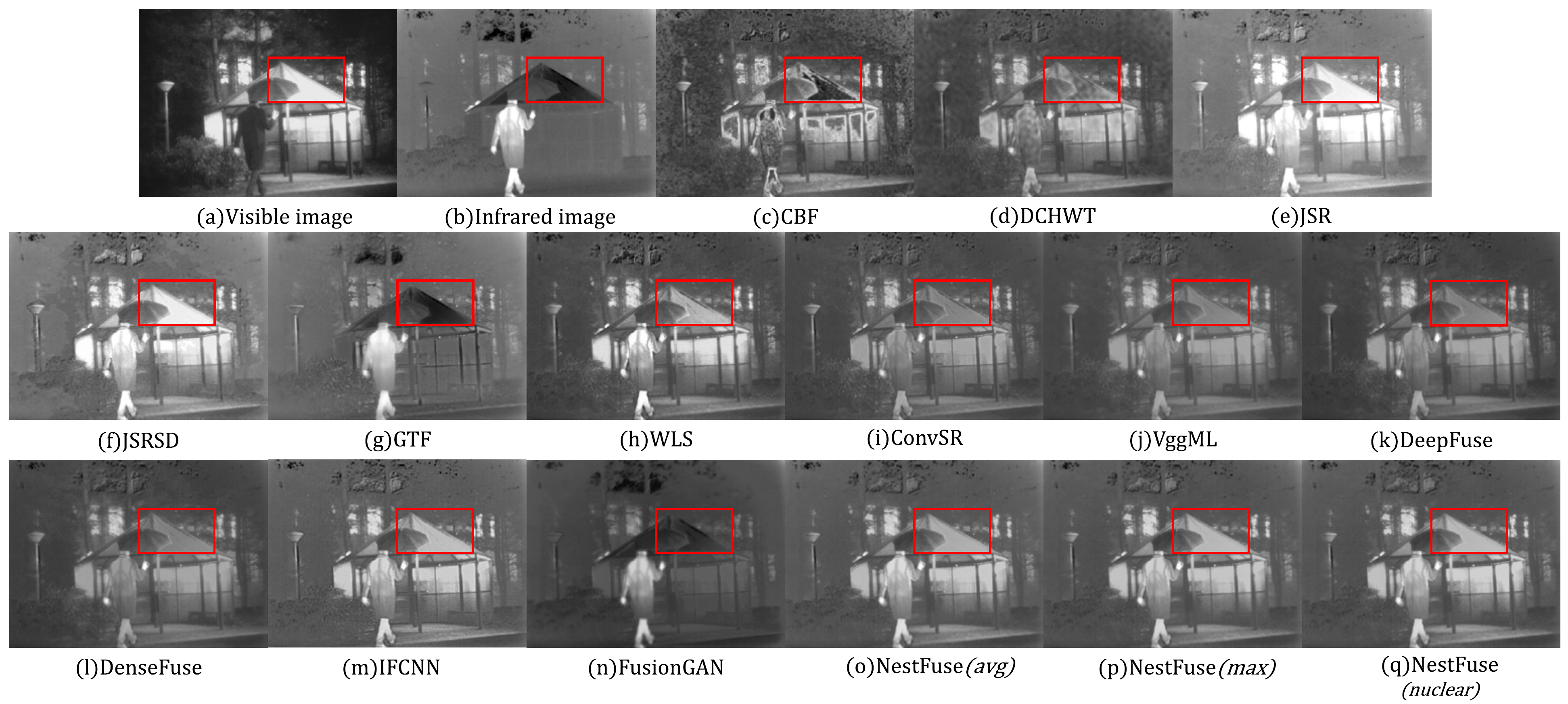}
\caption{Experiment on “umbrella” images. (a) Infrared image; (b) Visible image; (c) CBF; (d) DCHWT; (e) JSR; (f) JSRSD; (g) GTF; (h) WLS; (i) ConvSR; (j) VggML; (k) DeepFuse; (l) DenseFuse; (m) IFCNN; (n) FusionGAN; (o) NestFuse$(avg)$; (p) NestFuse$(max)$; (q) NestFuse$(nuclear)$.}
\label{fig:umbrella}
\end{figure*}

\begin{table*}[!ht]
\footnotesize
\centering
\caption{\label{tab:object} The average values of quality metrics for fused images. “$avg$”, “$max$” and “$nuclear$” denote the global pooling operator($P(\cdot)$) in channel attention model-based fusion strategy.}
\resizebox{0.9\linewidth}{!}{
\begin{tabular}{|c|c|c|c|c|c|c|c|c|}
\hline
 &  & $En$ \cite{roberts2008assessment} & SD \cite{rao1997fibre} & MI \cite{peng2005feature} & $FMI_{dct}$ \cite{haghighat2014fast} & $FMI_w$ \cite{haghighat2014fast} & $SSIM_a$ & VIF\cite{han2013new}\\
\hline
\multicolumn{2}{|c|}{CBF \cite{kumar2015image}}			&6.85749 	&76.82410 	&13.71498 	&0.26309 	&0.32350 	&0.59957 	&0.71849 \\
\hline
\multicolumn{2}{|c|}{DCHWT \cite{kumar2013multifocus}}			&6.56777 	&64.97891 	&13.13553 	&0.38568 	&0.40147 	&0.73132 	&0.50560 \\
\hline
\multicolumn{2}{|c|}{JSR \cite{zhang2013dictionary}}			&6.72263 	&74.10783 	&12.72654 	&0.14236 	&0.18506 	&0.60642 	&0.63845 \\
\hline
\multicolumn{2}{|c|}{JSRSD \cite{liu2017infrared}}			&6.72057 	&79.19536 	&13.38575 	&0.14253 	&0.18498 	&0.54097 	&0.67071 \\
\hline
\multicolumn{2}{|c|}{GTF \cite{ma2016infrared}}			&6.63433 	&67.54361 	&13.26865 	&0.39787 	&0.41038 	&0.70016 	&0.41687 \\
\hline
\multicolumn{2}{|c|}{WLS \cite{ma2017infrared}}			&6.64071 	&70.58894 	&13.28143 	&0.33103 	&0.37662 	&0.72360 	&0.72874 \\
\hline
\multicolumn{2}{|c|}{ConvSR \cite{liu2016image}}		&6.25869 	&50.74372 	&12.51737 	&0.34640 	&0.34640 	&\emph{\color{red}{0.75335}} 	&0.39218 \\
\hline
\multicolumn{2}{|c|}{VggML \cite{li2018infrared}}			&6.18260 	&48.15779 	&12.36521 	&0.40463 	&0.41684 	&\textbf{0.77803} 	&0.29509 \\
\hline
\multicolumn{2}{|c|}{DeepFuse \cite{ram2017deepfuse}}			&6.69935 	&68.79312 	&13.39869 	&\emph{\color{red}{0.41501}} 	&0.42477 	&0.72882 	&0.65773 \\
\hline
\multicolumn{2}{|c|}{DenseFuse \cite{li2018densefuse}}			&6.67158 	&67.57282 	&13.34317 	&\textbf{0.41727} 	&0.42767 	&0.73150 	&0.64576 \\
\hline
\multicolumn{2}{|c|}{FusionGAN \cite{ma2019fusiongan}}			&6.36285 	&54.35752 	&12.72570 	&0.36335 	&0.37083 	&0.65384 	&0.45355 \\
\hline 
\multicolumn{2}{|c|}{IFCNN \cite{zhang2020ifcnn}}			&6.59545 	&66.87578 	&13.19090 	&0.37378 	&0.40166 	&0.73186 	&0.59029 \\
\hline 

\multirow{3}*{NestFuse} &
$avg$		&\textbf{6.91971} 	&\emph{\color{red}{82.75242}} 	&\textbf{13.83942} 	&0.35801 	&\textbf{0.43724} 	&0.73199 	&\textbf{0.78652} \\
\cline{2-9}
~&$max$		&6.89421 	&80.36372 	&13.78842 	&0.36080 	&0.43293 	&0.73532 	&0.75204 \\
\cline{2-9}
~&$nuclear$	&\emph{\color{red}{6.90461}} 	&\textbf{82.92572} 	&\emph{\color{red}{13.80923}} 	&0.36277 	&\emph{\color{red}{0.43621}} 	&0.73360 	&\emph{\color{red}{0.76415}} \\
\hline

\end{tabular}}
\end{table*}

\subsubsection{\textbf{The Influence of Multi-scale Deep Features}}

In this section, we analyze the influence to fusion performance with different scales of deep features, the parameter $\lambda$ is set as 100.

To generate multiple outputs in different scales of deep features, we use deeply supervised training strategy which is utilized in UNet++ \cite{zhou2018unet++} to train our fusion network. The training framework of deep supervision of NestFuse is shown in Fig.\ref{fig:deeply-super}. 

$O_1$, $O_2$ and $O_3$ are outputs obtained by NestFuse with deep supervision. And the loss function $L$ is defined as follows,
\begin{eqnarray}\label{Eq12}
%  	L = \frac{1}{Q}\sum_{q=1}^Q(L_{pixel}(I,O_q) + \lambda L_{ssim}(I,O_q))
	L = \frac{1}{Q}\sum_{q=1}^Q(L_{total}(I,O_q))
\end{eqnarray}

\noindent where $Q=3$, $L_{total}$ is the total loss function which is discussed in section \ref{train-phase}.

%$L_{pixel}$ and $L_{ssim}$ are pixel loss function and SSIM loss function which were discussed in section \ref{train-phase}, and $\lambda$ is set as 1000.

Seven quality metrics are also selected to evaluate the fusion performance in different scales of deep features. These values are shown in Table \ref{tab:deep-super} and the best values are indicated in \textbf{bold}. ``w/o deep supervision'' denotes the training phase without deep supervision which was introduced in section \ref{train-phase}.

From Table \ref{tab:deep-super}, with deep supervision, the metrics values are very close in different scales ($O_1$, $O_2$ and $O_3$) and the advantage of multi-scale deep features in NestFuse is not competitive. Specifically, comparing with deep scale features (such as $O_3$), the shallow scale features ($O_2$) obtain better evaluation in $En$, $SD$, $MI$ and $VIF$, which indicates shallow scale features contain more detail information. When deeper scale features are utilized in NestFuse, the fused images contain more structure features, which delivers best values on $FMI_{dct}$, $SSIM_a$ and a comparable value on $FMI_w$.

However, when we train NestFuse with global optimization strategy, the fusion performance is boosted (obtains all best values), which means multi-scale mechanism is effective in our fusion network. This indicates that while the deeply supervised strategy may not train a better model in image fusion task, it still achieves better performance in image segmentation.

Thus, our network is trained with global optimization strategy which fully utilizes the multi-scale features in NestFuse.

%--------------------------------------------------------------------------------------------------
\subsection{Results Analysis}

The fused images obtained by existing fusion methods and our fusion method (NestFuse) are shown in Fig.\ref{fig:man} - Fig.\ref{fig:umbrella}. We analyze the visual effects of fused results on three pairs of infrared and visible images.

As shown in the red boxes of Fig.\ref{fig:man}, Fig.\ref{fig:street} and Fig.\ref{fig:umbrella}, comparing with the proposed method, CBF, DCHWT, JSR and JSRSD generate much more noise in fused images and some detail information are not clear. For GTF, WLS, ConvSR, VggML and FusionGAN, although some of the saliency features are highlighted, some regions in fused images are blurred. Moreover, the features in red boxes are not so satisfactory.

On the contrary, the DeepFuse, DenseFuse, IFCNN and the proposed method obtain better fusion performance in subjective evaluation compared with other three fusion methods. In addition, the fused images obtained by the proposed method have more reasonable luminance information.

For objective evaluation, we choose seven objective metrics to evaluate the fusion performance of these eleven fusion methods and the proposed method.

The average values of seven metrics for all fused images which are obtained by existing methods and the proposed fusion method are shown in Table \ref{tab:object}. The best values are indicated in \textbf{bold} and the second-best values are denoted in \emph{\color{red}{red and italic}}.

From Table \ref{tab:object}, the proposed fusion framework has five best values and five second-best values (except $FMI_{dct}$ and $SSIM_a$). This indicates that the proposed fusion framework can preserve more detail information ($En$, $SD$ and $MI$) and feature information($FMI_w$ and $VIF$) in the fused images.

For the metric $En$, it is used to measure the amount of information in one image. The larger $En$ means the fused image contains more information. However, if the fusion method generates noise in fuison processing, it also leads to larger $En$ (Fig.\ref{fig:man}(c)-\ref{fig:umbrella}(c)). That is why the fused image obtained by CBF achieves larger $En$. On the contrary, the fused images obtained by our proposed method have more reasonable luminance information and contain less noise, which make our proposed fusion method achieves best value on $En$.

Comparing with $max$ operator in channel attention model-based fusion strategy, the $nuclear$ and $average$ operators can achieve almost the best values in objective metrics. In channel attention model, these two operations are effective and can capture more structure information from deep features.

%--------------------------------------------------------------------------------------------------
\begin{figure*}[!ht]
\centering
\includegraphics[width=0.95\linewidth]{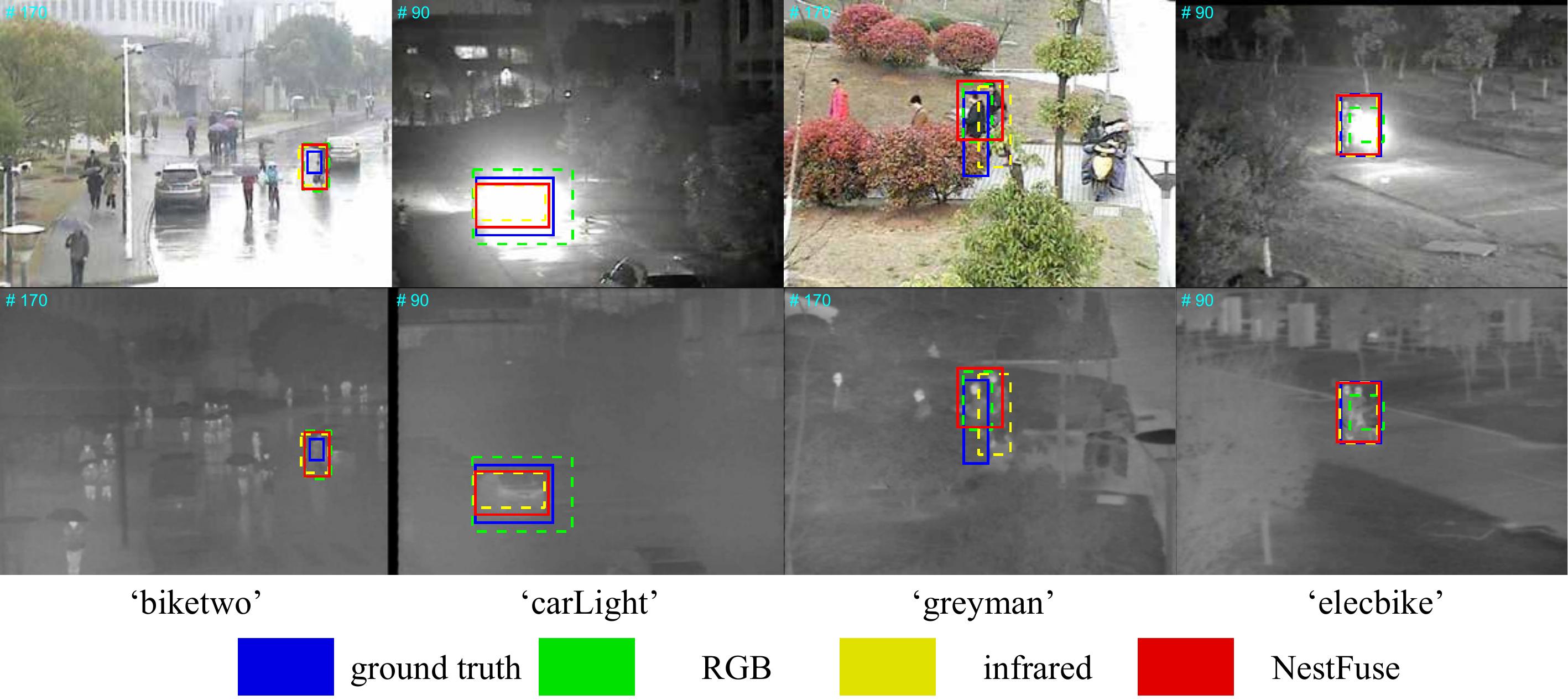}
\caption{The tracking results on the VOT-RGBT2019 benchmark. In first row and second row are RGB and infrared iamges, respectively. Each column includes one pair of RGB and infrared frames which are selected from video sequences `biketwo', `carLight', `greyman' and `elecbike'. The `RGB' and `infrared' denote the input of the SiamRPN++ is just one modality (RGB or infrared). The `NestFuse' presents the case when the multi-scale fusion strategy is applied into SiamRPN++.}
\label{fig:tracking}
\end{figure*}

\begin{table}[!ht]
\footnotesize
\centering
\caption{\label{tab:tracking} Tracking results using SiamRPN++ and NestFuse on VOT-RGBT2019 dataset. The fusion strategy which is developed in NestFuse is utilized to fuse multi-scale deep features.}
\resizebox{0.9\linewidth}{!}{
\begin{tabular}{|c|c|c|c|c|}
\hline
 & \emph{Type} & EAO & Accuracy & Failures  \\\hline
\multirow{2}*{SiamRPN++} &
\emph{infrared} & 0.2831 & 0.5875 & 43.9274  \\
\cline{2-5}
 & \emph{RGB} & \emph{\color{red}{0.3312}} & \emph{\color{red}{0.6104}} & \textbf{37.5201} \\\hline
\multicolumn{2}{|c|}{\emph{\makecell{SiamRPN++ \\with \textbf{NestFuse}}}} & \textbf{0.3493} & \textbf{0.6661} & \emph{\color{red}{40.9503}} \\\hline
\end{tabular}}
\end{table}

\subsection{An Application to Visual Objective Tracking}

The Visual Object Tracking (VOT) challenges address short-term or long-term, causal and model-free tracking \cite{vot2019rgbt}\cite{xu2019learning}\cite{xu2019joint}\cite{xu2019an}. 

In VOT2019, two new sub-challenges (VOT-RGBT and VOT-RGBD) are introduced by the committee. The VOT-RGBT sub-challenge focuses on short-term tracking, which contains two modalities (RGB and thermal infrared). As mentioned in \cite{li2020mdlatlrr}, the infrared and visible image fusion methods are ideally suited to improve the tracking performance in this task.

According to our previous research \cite{li2020mdlatlrr}, if the tracker engages a greater proportion of deep features for the data representation, its performance will be improved when the fusion method focuses on feature-level fusion. This insight gives us a direction to apply our proposed fusion method into RGBT tracking task. 

Thus, in this experiment, we choose SiamRPN++ \cite{li2019siamrpn++} as the base tracker and the fusion strategy proposed in this paper is applied to do the feature-level fusion. The SiamRPN++ is based on deep learning and achieves the state-of-the-art tracking performance in 2019.

In VOT-RGBT benchmark\cite{vot2019rgbt}, it contains 60 video sequences. The examples of these frames and some tracking results are shown in Fig.\ref{fig:tracking}. 

For the objective evaluation, three metrics \cite{vot2019rgbt} are selected to analyze the tracking performance: Expected Average Overlap (EAO), Accuracy and Failures. (1) EAO is an estimator of the average overlap a tracker manages to attain on a large collection with the same visual properties as the ground-truth; (2) Accuracy denotes the average overlap between the predicted and ground truth bounding boxes; (3) Failure evaluates the robustness of a tracker.

The evaluation measure values of SiamRPN++ with the proposed fusion method are shown in Table \ref{tab:tracking}. The \textbf{bold} and \emph{\color{red}{red italic}} indicate the best values and second-best values, respectively.

In VOT challenge, the EAO is the primary measure. As shown in Table \ref{tab:tracking}, comparing with `$RGB$' and `$infrared$', the tracking performance (EAO) is improved by applying our fusion strategy to fuse multi-scale deep features. This indicates that not only in image fusion task, the proposed fusion method can also improve the tracking performance in RGBT tracking task as well.

Furthermore, we will also apply the proposed fusion method into other computer vision tasks to evaluate the performance of fusion algorithm in future.

%-----------------------------------------------------------------------------------------------
\section{Conclusions}
\label{con}

In this paper, we propose a novel image fusion architecture by developing a nest connection network and spatial/channel attention models. Firstly, with the pooling operator in encoder network, the multi-scale features are extracted by this architecture, which could present richer features from source images. Then, the proposed spatial/channel attention models are utilized to fuse these multi-scale deep features in each scale. These fused features are fed into the nest connection-based decoder network to generate the fused image. With this novel network structure and the multi-scale deep feature fusion strategy, more saliency features can be preserved in the reconstruction process and the fusion performance can also be improved. 

The experimental results and analyses show that the proposed fusion framework demonstrates state-of-the-art fusion performance. An additional experiment on RGBT tracking task also shows that the proposed fusion strategy is effective in improving the algorithm performance in other computer vision task.

{\small
\bibliographystyle{unsrt}
\bibliography{nestfuse}

\begin{thebibliography}{10}

\bibitem{li2017pixel}
Shutao Li, Xudong Kang, Leyuan Fang, Jianwen Hu, and Haitao Yin.
\newblock {Pixel-level image fusion: A survey of the state of the art}.
\newblock {\em Information Fusion}, 33:100--112, 2017.

\bibitem{li2020mdlatlrr}
Hui Li, Xiao-Jun Wu, and Josef Kittler.
\newblock {MDLatLRR: A novel decomposition method for infrared and visible
  image fusion}.
\newblock {\em IEEE Transactions on Image Processing}, 2020.
\newblock doi: 10.1109/TIP.2020.2975984.

\bibitem{li2019rgb}
Chenglong Li, Xinyan Liang, Yijuan Lu, Nan Zhao, and Jin Tang.
\newblock {RGB-T object tracking: benchmark and baseline}.
\newblock {\em Pattern Recognition}, 96:106977, 2019.

\bibitem{vot2019rgbt}
Matej Kristan, Jiri Matas, Ales Leonardis, Michael Felsberg, Roman Pflugfelder,
  Joni-Kristian Kamarainen, Luka Cehovin~Zajc, Ondrej Drbohlav, Alan Lukezic,
  Amanda Berg, et~al.
\newblock {The seventh visual object tracking vot2019 challenge results}.
\newblock In {\em Proceedings of the IEEE International Conference on Computer
  Vision Workshops}, pages 1--36, 2019.

\bibitem{pajares2004wavelet}
Gonzalo Pajares and Jesus~Manuel De~La~Cruz.
\newblock {A wavelet-based image fusion tutorial}.
\newblock {\em Pattern recognition}, 37(9):1855--1872, 2004.

\bibitem{ben2005multiscale}
A~Ben~Hamza, Yun He, Hamid Krim, and Alan Willsky.
\newblock {A multiscale approach to pixel-level image fusion}.
\newblock {\em Integrated Computer-Aided Engineering}, 12(2):135--146, 2005.

\bibitem{li2020laplacian}
Xiaoxiao Li, Xiaopeng Guo, Pengfei Han, Xiang Wang, Huaguang Li, and Tao Luo.
\newblock Laplacian re-decomposition for multimodal medical image fusion.
\newblock {\em IEEE Transactions on Instrumentation and Measurement}, 2020.
\newblock doi: 10.1109/TIM.2020.2975405.

\bibitem{liu2017infrared}
CH~Liu, Y~Qi, and WR~Ding.
\newblock {Infrared and visible image fusion method based on saliency detection
  in sparse domain}.
\newblock {\em Infrared Physics \& Technology}, 83:94--102, 2017.

\bibitem{li2019discriminative}
Huafeng Li, Yitang Wang, Zhao Yang, Ruxin Wang, Xiang Li, and Dapeng Tao.
\newblock Discriminative dictionary learning-based multiple component
  decomposition for detail-preserving noisy image fusion.
\newblock {\em IEEE Transactions on Instrumentation and Measurement}, 2019.
\newblock doi: 10.1109/TIM.2019.2912239.

\bibitem{li2017multi}
Hui Li and Xiao-Jun Wu.
\newblock {Multi-focus image fusion using dictionary learning and low-rank
  representation}.
\newblock In {\em International Conference on Image and Graphics}, pages
  675--686. Cham, Switzerland: Springer, 2017.

\bibitem{liu2016image}
Yu~Liu, Xun Chen, Rabab~K Ward, and Z~Jane Wang.
\newblock Image fusion with convolutional sparse representation.
\newblock {\em IEEE signal processing letters}, 23(12):1882--1886, 2016.

\bibitem{li2018infrared}
Hui Li, Xiao-Jun Wu, and Josef Kittler.
\newblock {Infrared and Visible Image Fusion using a Deep Learning Framework}.
\newblock In {\em 2018 24th International Conference on Pattern Recognition
  (ICPR)}, pages 2705--2710. IEEE, 2018.

\bibitem{li2019infrared}
Hui Li, Xiao-Jun Wu, and Tariq~S Durrani.
\newblock {Infrared and Visible Image Fusion with ResNet and zero-phase
  component analysis}.
\newblock {\em Infrared Physics \& Technology}, page 103039, 2019.

\bibitem{li2018densefuse}
Hui Li and Xiao-Jun Wu.
\newblock {DenseFuse: A Fusion Approach to Infrared and Visible Images}.
\newblock {\em IEEE Transactions on Image Processing}, 28(5):2614--2623, 2018.

\bibitem{yang2010image}
Shuyuan Yang, Min Wang, Licheng Jiao, Ruixia Wu, and Zhaoxia Wang.
\newblock {Image fusion based on a new contourlet packet}.
\newblock {\em Information Fusion}, 11(2):78--84, 2010.

\bibitem{li2013image}
Shutao Li, Xudong Kang, and Jianwen Hu.
\newblock {Image fusion with guided filtering}.
\newblock {\em IEEE Transactions on Image processing}, 22(7):2864--2875, 2013.

\bibitem{vishwakarma2018image}
Amit Vishwakarma and MK~Bhuyan.
\newblock Image fusion using adjustable non-subsampled shearlet transform.
\newblock {\em IEEE Transactions on Instrumentation and Measurement},
  68(9):3367--3378, 2018.

\bibitem{wright2008robust}
John Wright, Allen~Y Yang, Arvind Ganesh, S~Shankar Sastry, and Yi~Ma.
\newblock Robust face recognition via sparse representation.
\newblock {\em IEEE transactions on pattern analysis and machine intelligence},
  31(2):210--227, 2008.

\bibitem{liu2010robust}
Guangcan Liu, Zhouchen Lin, and Yong Yu.
\newblock Robust subspace segmentation by low-rank representation.
\newblock In {\em ICML}, volume~1, page~8, 2010.

\bibitem{singh2019multimodal}
Sneha Singh and RS~Anand.
\newblock Multimodal medical image sensor fusion model using sparse k-svd
  dictionary learning in nonsubsampled shearlet domain.
\newblock {\em IEEE Transactions on Instrumentation and Measurement}, 2019.
\newblock doi: 10.1109/TIM.2019.2902808.

\bibitem{aharon2006k}
Michal Aharon, Michael Elad, Alfred Bruckstein, et~al.
\newblock {K-SVD: An algorithm for designing overcomplete dictionaries for
  sparse representation}.
\newblock {\em IEEE Transactions on signal processing}, 54(11):4311, 2006.

\bibitem{lu2014infrared}
Xiaoqi Lu, Baohua Zhang, Ying Zhao, He~Liu, and Haiquan Pei.
\newblock {The infrared and visible image fusion algorithm based on target
  separation and sparse representation}.
\newblock {\em Infrared Physics \& Technology}, 67:397--407, 2014.

\bibitem{yin2017novel}
Ming Yin, Puhong Duan, Wei Liu, and Xiangyu Liang.
\newblock {A novel infrared and visible image fusion algorithm based on
  shift-invariant dual-tree complex shearlet transform and sparse
  representation}.
\newblock {\em Neurocomputing}, 226:182--191, 2017.

\bibitem{liu2017multi}
Yu~Liu, Xun Chen, Hu~Peng, and Zengfu Wang.
\newblock {Multi-focus image fusion with a deep convolutional neural network}.
\newblock {\em Information Fusion}, 36:191--207, 2017.

\bibitem{yan2018unsupervised}
Xiang Yan, Syed~Zulqarnain Gilani, Hanlin Qin, and Ajmal Mian.
\newblock Unsupervised deep multi-focus image fusion.
\newblock {\em arXiv preprint arXiv:1806.07272}, 2018.

\bibitem{ma2019fusiongan}
Jiayi Ma, Wei Yu, Pengwei Liang, Chang Li, and Junjun Jiang.
\newblock {FusionGAN: A generative adversarial network for infrared and visible
  image fusion}.
\newblock {\em Information Fusion}, 48:11--26, 2019.

\bibitem{zhang2020ifcnn}
Yu~Zhang, Yu~Liu, Peng Sun, Han Yan, Xiaolin Zhao, and Li~Zhang.
\newblock {IFCNN: A general image fusion framework based on convolutional
  neural network}.
\newblock {\em Information Fusion}, 54:99--118, 2020.

\bibitem{simonyan2014very}
Karen Simonyan and Andrew Zisserman.
\newblock {Very deep convolutional networks for large-scale image recognition}.
\newblock {\em arXiv preprint arXiv:1409.1556}, 2014.

\bibitem{he2016deep}
Kaiming He, Xiangyu Zhang, Shaoqing Ren, and Jian Sun.
\newblock Deep residual learning for image recognition.
\newblock In {\em Proceedings of the IEEE conference on computer vision and
  pattern recognition}, pages 770--778, 2016.

\bibitem{krizhevsky2012imagenet}
Alex Krizhevsky, Ilya Sutskever, and Geoffrey~E Hinton.
\newblock Imagenet classification with deep convolutional neural networks.
\newblock In {\em Advances in neural information processing systems}, pages
  1097--1105, 2012.

\bibitem{huang2017densely}
Gao Huang, Zhuang Liu, Laurens Van Der~Maaten, and Kilian~Q Weinberger.
\newblock Densely connected convolutional networks.
\newblock In {\em Proceedings of the IEEE conference on computer vision and
  pattern recognition}, pages 4700--4708, 2017.

\bibitem{goodfellow2014generative}
Ian Goodfellow, Jean Pouget-Abadie, Mehdi Mirza, Bing Xu, David Warde-Farley,
  Sherjil Ozair, Aaron Courville, and Yoshua Bengio.
\newblock Generative adversarial nets.
\newblock In {\em Advances in neural information processing systems}, pages
  2672--2680, 2014.

\bibitem{zhou2018unet++}
Zongwei Zhou, Md~Mahfuzur~Rahman Siddiquee, Nima Tajbakhsh, and Jianming Liang.
\newblock {Unet++: A nested u-net architecture for medical image segmentation}.
\newblock In {\em Deep Learning in Medical Image Analysis and Multimodal
  Learning for Clinical Decision Support}, pages 3--11. Springer, 2018.

\bibitem{liu2018deep}
Yu~Liu, Xun Chen, Zengfu Wang, Z~Jane Wang, Rabab~K Ward, and Xuesong Wang.
\newblock Deep learning for pixel-level image fusion: Recent advances and
  future prospects.
\newblock {\em Information Fusion}, 42:158--173, 2018.

\bibitem{wang2004image}
Zhou Wang, Alan~C Bovik, Hamid~R Sheikh, Eero~P Simoncelli, et~al.
\newblock Image quality assessment: from error visibility to structural
  similarity.
\newblock {\em IEEE transactions on image processing}, 13(4):600--612, 2004.

\bibitem{lin2014microsoft}
Tsung-Yi Lin, Michael Maire, Serge Belongie, James Hays, Pietro Perona, Deva
  Ramanan, Piotr Doll{\'a}r, and C~Lawrence Zitnick.
\newblock {Microsoft coco: Common objects in context}.
\newblock In {\em European conference on computer vision}, pages 740--755.
  Springer, 2014.

\bibitem{ma2017infrared}
Jinlei Ma, Zhiqiang Zhou, Bo~Wang, and Hua Zong.
\newblock Infrared and visible image fusion based on visual saliency map and
  weighted least square optimization.
\newblock {\em Infrared Physics \& Technology}, 82:8--17, 2017.

\bibitem{tno2018}
Alexander Toet.
\newblock {TNO Image Fusion Dataset}, 2014.
\newblock \url{https://figshare.com/articles/TN_Image_Fusion_Dataset/1008029}.

\bibitem{kumar2015image}
BK~Shreyamsha Kumar.
\newblock Image fusion based on pixel significance using cross bilateral
  filter.
\newblock {\em Signal, image and video processing}, 9(5):1193--1204, 2015.

\bibitem{kumar2013multifocus}
BK~Shreyamsha Kumar.
\newblock Multifocus and multispectral image fusion based on pixel significance
  using discrete cosine harmonic wavelet transform.
\newblock {\em Signal, Image and Video Processing}, 7(6):1125--1143, 2013.

\bibitem{zhang2013dictionary}
Qiheng Zhang, Yuli Fu, Haifeng Li, and Jian Zou.
\newblock Dictionary learning method for joint sparse representation-based
  image fusion.
\newblock {\em Optical Engineering}, 52(5):057006, 2013.

\bibitem{ma2016infrared}
Jiayi Ma, Chen Chen, Chang Li, and Jun Huang.
\newblock Infrared and visible image fusion via gradient transfer and total
  variation minimization.
\newblock {\em Information Fusion}, 31:100--109, 2016.

\bibitem{ram2017deepfuse}
K~Ram~Prabhakar, V~Sai~Srikar, and R~Venkatesh~Babu.
\newblock Deepfuse: a deep unsupervised approach for exposure fusion with
  extreme exposure image pairs.
\newblock In {\em Proceedings of the IEEE International Conference on Computer
  Vision}, pages 4714--4722, 2017.

\bibitem{roberts2008assessment}
J~Wesley Roberts, Jan~A Van~Aardt, and Fethi~Babikker Ahmed.
\newblock Assessment of image fusion procedures using entropy, image quality,
  and multispectral classification.
\newblock {\em Journal of Applied Remote Sensing}, 2(1):023522, 2008.

\bibitem{rao1997fibre}
Yun-Jiang Rao.
\newblock In-fibre bragg grating sensors.
\newblock {\em Measurement science and technology}, 8(4):355, 1997.

\bibitem{peng2005feature}
Hanchuan Peng, Fuhui Long, and Chris Ding.
\newblock Feature selection based on mutual information: criteria of
  max-dependency, max-relevance, and min-redundancy.
\newblock {\em IEEE Transactions on Pattern Analysis \& Machine Intelligence},
  (8):1226--1238, 2005.

\bibitem{haghighat2014fast}
Mohammad Haghighat and Masoud~Amirkabiri Razian.
\newblock {Fast-FMI: non-reference image fusion metric}.
\newblock In {\em 2014 IEEE 8th International Conference on Application of
  Information and Communication Technologies (AICT)}, pages 1--3. IEEE, 2014.

\bibitem{han2013new}
Yu~Han, Yunze Cai, Yin Cao, and Xiaoming Xu.
\newblock A new image fusion performance metric based on visual information
  fidelity.
\newblock {\em Information fusion}, 14(2):127--135, 2013.

\bibitem{xu2019learning}
Tianyang Xu, Zhen-Hua Feng, Xiao-Jun Wu, and Josef Kittler.
\newblock {Learning Adaptive Discriminative Correlation Filters via Temporal
  Consistency preserving Spatial Feature Selection for Robust Visual Object
  Tracking}.
\newblock {\em IEEE Transactions on Image Processing}, 2019.

\bibitem{xu2019joint}
Tianyang Xu, Zhen-Hua Feng, Xiao-Jun Wu, and Josef Kittler.
\newblock {Joint group feature selection and discriminative filter learning for
  robust visual object tracking}.
\newblock In {\em Proceedings of the IEEE International Conference on Computer
  Vision}, pages 7950--7960, 2019.

\bibitem{xu2019an}
Tianyang Xu, Zhen-Hua Feng, Xiao-Jun Wu, and Josef Kittler.
\newblock {An Accelerated Correlation Filter Tracker}.
\newblock {\em Pattern Recognition accepted. arXiv preprint arXiv:1912.02854},
  2019.

\bibitem{li2019siamrpn++}
Bo~Li, Wei Wu, Qiang Wang, Fangyi Zhang, Junliang Xing, and Junjie Yan.
\newblock {SiamRPN++: Evolution of siamese visual tracking with very deep
  networks}.
\newblock In {\em Proceedings of the IEEE Conference on Computer Vision and
  Pattern Recognition}, pages 4282--4291, 2019.

\end{thebibliography}
}

\begin{IEEEbiography}[{\includegraphics[width=1in,height=1.25in,clip,keepaspectratio]{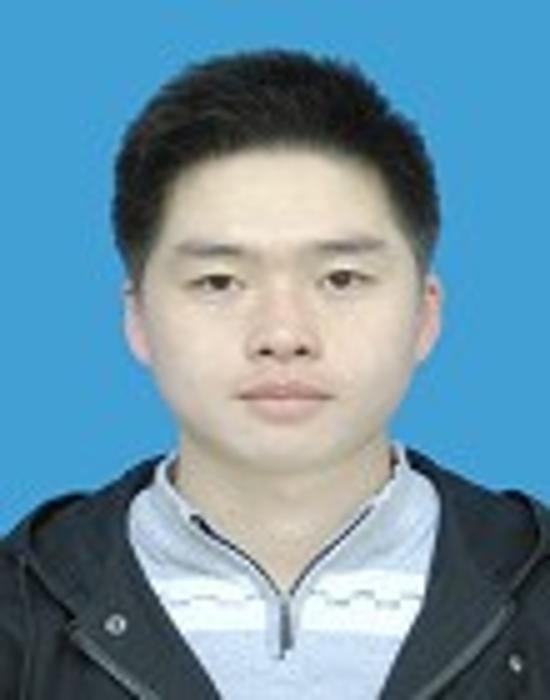}}]{Hui Li}
received the B.Sc. degree in School of Internet of Things Engineering from Jiangnan University, China, in 2015. He is currently a PhD student at the Jiangsu Provincial Engineerinig Laboratory of Pattern Recognition and Computational Intelligence, School of Artificial Intelligence and Computer Science, Jiangnan University. His research interests include image fusion, machine learning and deep learning.
\end{IEEEbiography}

\begin{IEEEbiography}[{\includegraphics[width=1in,height=1.25in,clip,keepaspectratio]{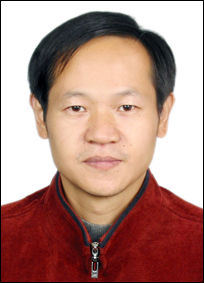}}]{Xiao-Jun Wu}
received the B.Sc. degree in mathematics from Nanjing Normal University, Nanjing, China,
in 1991, and the M.S. degree and Ph.D. degree
in pattern recognition and intelligent system from
the Nanjing University of Science and Technology, Nanjing, in 1996 and 2002, respectively. From
1996 to 2006, he taught at the School of Electronics
and Information, Jiangsu University of Science and
Technology, where he was promoted to Professor.

He has been with the School of Information Engineering, Jiangnan University since 2006, where he
is a Professor of pattern recognition and computational intelligence. He was a Visiting
Researcher with the Centre for Vision, Speech, and Signal Processing (CVSSP), University of Surrey, U.K. from 2003 to 2004. He has published over 300 papers in his fields of research.
His current research interests include pattern recognition, computer vision, and
computational intelligence. He was a Fellow of the International Institute for
Software Technology, United Nations University, from 1999 to 2000. He was
a recipient of the Most Outstanding Postgraduate Award from the Nanjing
University of Science and Technology.
\end{IEEEbiography}

\begin{IEEEbiography}[{\includegraphics[width=1in,height=1.25in,clip,keepaspectratio]{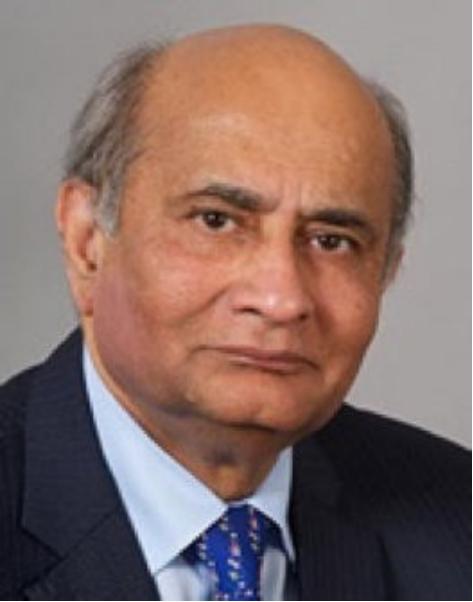}}]{Tariq Durrani}
is Research Professor at University of Strathclyde, Glasgow Scotland. His research covers AI, Signal Processing and Technology Management. He has authored 350 ublications; supervised 45 PhDs.
He is a Fellow of the: IEEE, UK Royal Academy of Engineering, Royal Society of Edinburgh, IET, and the Third World Academy of Sciences. In 2018 he was elected Foreign Member of the US National Academy of Engineering.
\end{IEEEbiography}

\end{document}